\documentclass[10pt,twocolumn,letterpaper]{article}

\usepackage{cvpr}
\usepackage{times}
\usepackage{epsfig}
\usepackage{graphicx}
\usepackage{amsmath}
\usepackage{amssymb}
\usepackage{multirow}
\usepackage[]{algorithm2e}
\usepackage[breaklinks=true,bookmarks=false]{hyperref}

\cvprfinalcopy % *** Uncomment this line for the final submission

\def\cvprPaperID{****} % *** Enter the CVPR Paper ID here

\begin{document}

%%%%%%%%% TITLE
\title{On zero-shot recognition of generic objects}

\author{Tristan Hascoet\\
{\tt\small tristan.hascoet@gmail.com}
\and
Yasuo Ariki\\
{\tt\small ariki@kobe-u.ac.jp}
\and
Tetsuya Takiguchi\\
{\tt\small takigu@kobe-u.ac.jp}
\and
Graduate School of System Informatics, Kobe University, Japan\\
}

\maketitle
%\thispagestyle{empty}

%%%%%%%%% ABSTRACT
\begin{abstract}
Many recent advances in computer vision are the result of a healthy competition among researchers on high quality, task-specific, benchmarks.
After a decade of active research, zero-shot learning (ZSL) models accuracy on the Imagenet benchmark remains far too low to be considered for practical object recognition applications.
In this paper, we argue that the main reason behind this apparent lack of progress is the poor quality of this benchmark.
We highlight major structural flaws of the current benchmark and analyze different factors impacting the accuracy of ZSL models.
We show that the actual classification accuracy of existing ZSL models is significantly higher than was previously thought as we account for these flaws. 
We then introduce the notion of structural bias specific to ZSL datasets. 
We discuss how the presence of this new form of bias allows for a trivial solution to the standard benchmark and conclude on the need for a new benchmark.
We then detail the semi-automated construction of a new benchmark to address these flaws.
\end{abstract}

%%%%%%%%% BODY TEXT
\section{Introduction}

% Leading roles of datasets in research
Datasets play a leading role in computer vision research.
Perhaps the most striking example of the impact a dataset can have on research has been the introduction of Imagenet \cite{deng2009imagenet}.
The new scale and granularity of Imagenet's coverage of the visual world has paved the way for the success and wide spread adoption of 
CNN \cite{krizhevsky2012imagenet,lecun1998gradient} that have revolutionized generic object recognition.

% Goals & promises of ZSL
The current best-practice for the development of a practical object recognition solution consists in 
collecting and annotating application-specific training data to fine-tune a large Imagenet-pretrained CNN on. 
This data annotation process can be prohibitively expensive for many applications which hinders the wide-spread usage of these technologies.
ZSL models generalize the recognition ability of traditional image classifiers to unknown classes, for which no image sample is available for training.
The promise of ZSL for generic object recognition is huge: 
to scale up the recognition capacity of image classifiers beyond the set of annotated training classes.
Hence ZSL has the potential to be of great practical impact as they would considerably ease the deployment of object recognition technologies
by eliminating the need for expensive task-specific data collection and fine-tuning processes.
%to eliminate the need for expensive task-specific data collection and fine-tuning in the process of deploying a practical object recognition solution.
%Successful ZSL models would be of great practical impact as they would considerably ease the deployment of practical object recognition solutions, accelerating their wide-spread adoption.

% Flaws of the current dataset
Despite its great promise, and after a decade of active research \cite{larochelle2008zero}, 
the accuracy of ZSL models on the standard Imagenet benchmark \cite{frome2013devise} remain far too low for practical applications.
To better understand this lack of progress, we analyzed the errors of several ZSL baselines.
Our analysis leads us to identify two main factors impacting the accuracy of ZSL models:
structural flaws in the standard evaluation protocol and poor quality of both semantic and visual samples.
On the bright side of things, we show that once these flaws are taken into account, 
the actual accuracy of existing ZSL models is much higher than was previously thought.

% Structural bias and compositional reasoning
On the other hand, we show that a trivial solution outperforms most existing ZSL models by a large margin, which is upsetting.
To explain this phenomenon, we introduce the notion of structural bias in ZSL datasets.
We argue that ZSL models should aim to develop compositional reasoning abilities, 
but the presence of structural bias in the Imagenet benchmark favors solutions based on a trivial one to one mapping between training and test classes.
We come to the conclusion that a new benchmark is needed to address the different problems identified by our analysis and, %in the current ZSL benchmark.
in the last section of this paper, we detail the semi-automated construction of a new benchmark we propose. % addressing these problems.

%Paper summary
To structure our discussion, we first briefly review preliminaries on ZSL in Section 3.
Section 4 details our analysis of the different factors impacting the accuracy of ZSL models on the standard benchmark.
Section 5 introduces the notions of structural bias, 
and propose a way to measure and minimize its impact in the construction of a new benchmark.
Finally, Section 6 summarizes the construction of our proposed benchmark.
For space constraint, we only include the main results of our analysis in the body of this paper. 
We refer interested readers to the supplementary material for additional results and details of our analysis.

%-------------------------------------------------------------------------
\section{Related Work}
\subsection{ZSL datasets}

% Small scale benchmarks
Early research on ZSL has been carried out on relatively small scale or domain specific benchmarks \cite{lampert2009learning,patterson2012sun,welinder2010caltech}, 
for which human-annotated visual attributes are proposed as semantic representations of the visual classes. 
On the one hand, these benchmarks have provided a controlled setup for the development of theoretical models and the accurate tracking of ZSL progress. 
On the other hand, it is unclear whether approaches developed on such dataset would generalize to the more practical setting of zero-shot generic object recognition.
For instance, in generic object recognition, manually annotating each and every possible visual class of interest with a set of visual attributes is impractical due to the diversity and complexity of the visual world. 
%Hence, generalizing the zero-shot learning approaches developed on such benchmarks to the more practical case of generic object recognition comes with the additional challenge of collecting suitable descriptions of the visual classes.

% ImageNet & DeViSe
The Imagenet dataset \cite{deng2009imagenet} consists of more than 13 million images scattered among 21,845 visual classes.
Imagenet relies on Wordnet \cite{miller1995wordnet} to structure its classes: each visual class in Imagenet corresponds to a concept in Wordnet.
Frome \etal \cite{frome2013devise} proposed a benchmark for ZS generic object recognition based on the Imagenet dataset,
which has been widely adopted as the standard evaluation benchmark by recent works
\cite{norouzi2013zero,xian2016latent,romera2015embarrassingly,changpinyo2016synthesized,xian2017zero,kampffmeyer2018rethinking,wang2018zero}.
Using word embeddings as semantic representations, 
they use the 1000 classes of the ILSVRC dataset as training classes
and propose different test splits drawn from the remaining 20,845 classes of the Imagenet dataset 
based on their distance to the training classes within the Wordnet hierarchy:
the \textit{2-hops}, \textit{3-hops} and \textit{all} test splits.

% Flaws in DeViSe
Careful inspection of these test splits revealed a confusion in their name: 
The \textit{2-hops} test split actually consists of the set of $1589$ test classes directly connected to the training set classes in Wordnet, i.e; within \textit{1 hop} of the training set.
Similarly, the \textit{3-hops} test set actually corresponds to the test classes within \textit{2-hops}.
In this paper, we will refer to the standard test splits by the name of their true configuration:
\textit{1-hop}, \textit{2-hops} and \textit{all}, as illustrated in Figure 1.

\subsection{Dataset bias}

%Generic object recognition
Bias in datasets can take many forms, depending on the specific target task.
Torralba \etal \cite{torralba2011unbiased} investigates bias in generic object recognition. 
The notion of structural bias we introduce in Section 5 is closely related to the notion of negative set bias they analyze.
%which we will relate to the notion of structural bias in ZSL.

% VQA
As more complex tasks are being considered, more insidious forms of bias sneak into our datasets.
In VQA, the impressive results of early baseline models
have later been shown to be largely due to statistical biases in the question/answers pairs \cite{jabri2016revisiting,kafle2017analysis,johnson2017clevr}.
Similar to these works, we will show that a trivial solution leveraging structural bias in the Imagenet ZSL benchmark outperforms early ZSL baselines.

% ZSL
Xian \etal \cite{xian2017zero} identify structural incoherences in small-scale ZSL benchmarks and proposes new test splits to remedy them.
Closely related to our work, they also observe a correlation between test class sample population and classification accuracy in the Imagenet ZSL benchmark.
However, their analysis mainly focuses on small-scale benchmarks and the comparison of existing ZSL models, 
while we analyze the ZSL benchmark for generic object recognition in more depth.

\section{Preliminaries}

ZSL models aim to recognize unseen classes, for which no image sample is available to learn from. 
To do so, ZSL models use descriptions of the visual classes, i.e., representations of the visual classes in a semantic space shared by both training and test classes.
To evaluate the out-of-sample recognition ability of models, ZSL benchmarks split the full set of classes $C$ into disjoint training and test sets.
ZSL benchmarks are fully defined by three components:
a set of training and test classes $(C_{tr}, C_{te})$, 
a set of labeled images $X$, and a set of semantic representations $Y$:% of each visual class.

\begin{subequations} 
\begin{align}
& C_{tr} \cup C_{te} \subset C         \\
& C_{tr} \cap C_{te} = \emptyset       \\
& Y = \{y_c \in \mathbb{R}^d \quad \forall c \in C\} \\
& X = \{(x,c) \in \mathbb{R}^{3 \times h \times w} \times C\}   \\
& Tr = \{(x,y_c) \; | \; c \in C_{tr}\}  \\
& Te = \{(x,y_c) \; | \; c \in C_{te}\} 
\end{align}
\end{subequations} 

%\subsection{ZSL models}
ZSL models are typically trained to minimize a loss function $\mathcal{L}$ over a similarity score $E$ between image and semantic features of the training sample set with respect to the model parameters $\theta$.  
\begin{equation} 
\theta^{*} = argmin_{\theta} \mathbb{E}_{(x,y) \in Tr}\mathcal{L}(E_{\theta}(x,y) + \Omega(\theta))
\end{equation} 
In the standard ZSL setting, test samples $x_{te}$ are classified among the set of unseen test classes by retrieving the class description $y$ of highest similarity score:
\begin{equation} 
c = argmax_{c \in C_{te}}E(x_{te}, y_c)
\end{equation} 
In the generalized ZSL setting, test samples are classified among the full set of training and test classes:
\begin{equation} 
c = argmax_{c \in C}E(x_{te}, y_c)
\end{equation} 

% Brief history of models
Xian \etal \cite{xian2016latent} have shown that many ZSL models can be formulated within a same linear model framework, 
with different training objectives and regularization terms.
More recently, Wang \etal \cite{wang2018zero} have proposed a Graph Convolutional Network (GCN) 
model that has shown impressive improvements over the previous state of the art.
In our study, we will present results obtained with both a baseline linear model \cite{romera2015embarrassingly} 
and a state of the art GCN model \cite{wang2018zero,kampffmeyer2018rethinking}.

\section{Error analysis}

% Prerequisite
%In the previous section, we have mentioned that ZSL benchmarks are fully defined by three components: $X$, $Y$ and $(C_{tr}, C_{te})$.
%ZSL benchmarks should only reflect the ability of models to perform ZSL recognition.
%To do so, each of these three components should be unambiguous to allow for such a solution to exist.
%The reported accuracy should not reflect in any way an ambiguous 
%The goal of a benchmark is to provide an unambiguous evaluation protocol to fairly track progress and gains in ZSL accuracy.
%To do so, 
%In this section, we argue that in order to fairly evaluate ZSL models, 
%a set of labeled images $X$, a set of semantic representations $Y$, and the set of training and test classes: $(C_{tr}, C_{te})$.
%We argue that a good ZSL benchmark should only evaluate the zero-shot learning recognition of models.
%To do so, it should provide unambiguous X, Y and D.
In the previous section, we have mentioned that ZSL benchmarks are fully defined by three components: %$X$, $Y$ and $(C_{tr}, C_{te})$.
a set of labeled images $X$, a set of semantic representations $Y$, and the set of training and test classes $(C_{tr}, C_{te})$.
In this section, we analyze each of the standard benchmark components individually:
We first highlight inconsistencies in the configuration of the different test splits 
and show that these inconsistencies lead to many false negatives in the reported evaluation of ZSL models outputs.
Next, we identify a number of factors impacting the quality of the word embeddings 
of visual classes and argue that visual classes with poor semantic representations should be excluded from ZSL benchmarks.
We then observe that the Imagenet dataset contains many ambiguous image samples.
We define what a \textit{good} image sample means in the context of ZSL
and propose a method to automatically select such images.

\subsection{Structural flaws}

% Problem.
Figure 1 illustrates the configuration of test classes of the standard test splits within the Wordnet hierarchy.
This configuration leads to an obvious contradiction: test sets include visual classes of both parents and their children concepts.
Consider the problem of classifying images of birds within the \textit{hop-1} test split as in Figure 1.
The standard test splits give rise to two possibly inconsistent scenarios:

\begin{figure}[h]
\centering
\includegraphics[width=0.35\textwidth]{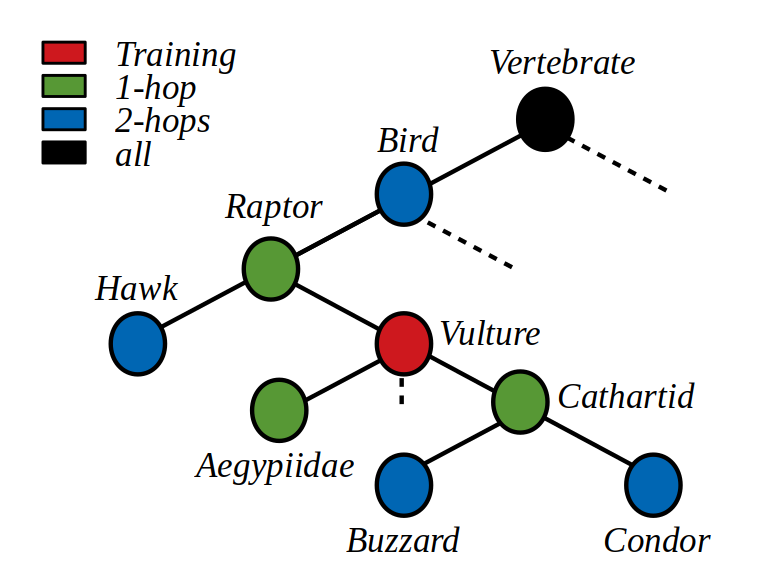}
\caption{Illustration of the standard test splits configuration}
\end{figure}

A ZSL model may classify an image of the children class \textit{Cathartid} as its parent class \textit{Raptor}.
The standard benchmark considers such cases as classification errors, while the classification is semantically correct.

A ZSL model may classify an image of the parent class \textit{Raptor} as one of its children class: \textit{Cathartid}.
Classification may be semantically correct or incorrect, 
depending on the specific breed of raptor in the image, 
but we have no way to automatically assess it without additional annotation.
The standard benchmark considers such cases as classification errors, while the classification is semantically undefined.

\begin{figure}[h]
\centering
\includegraphics[width=0.4\textwidth]{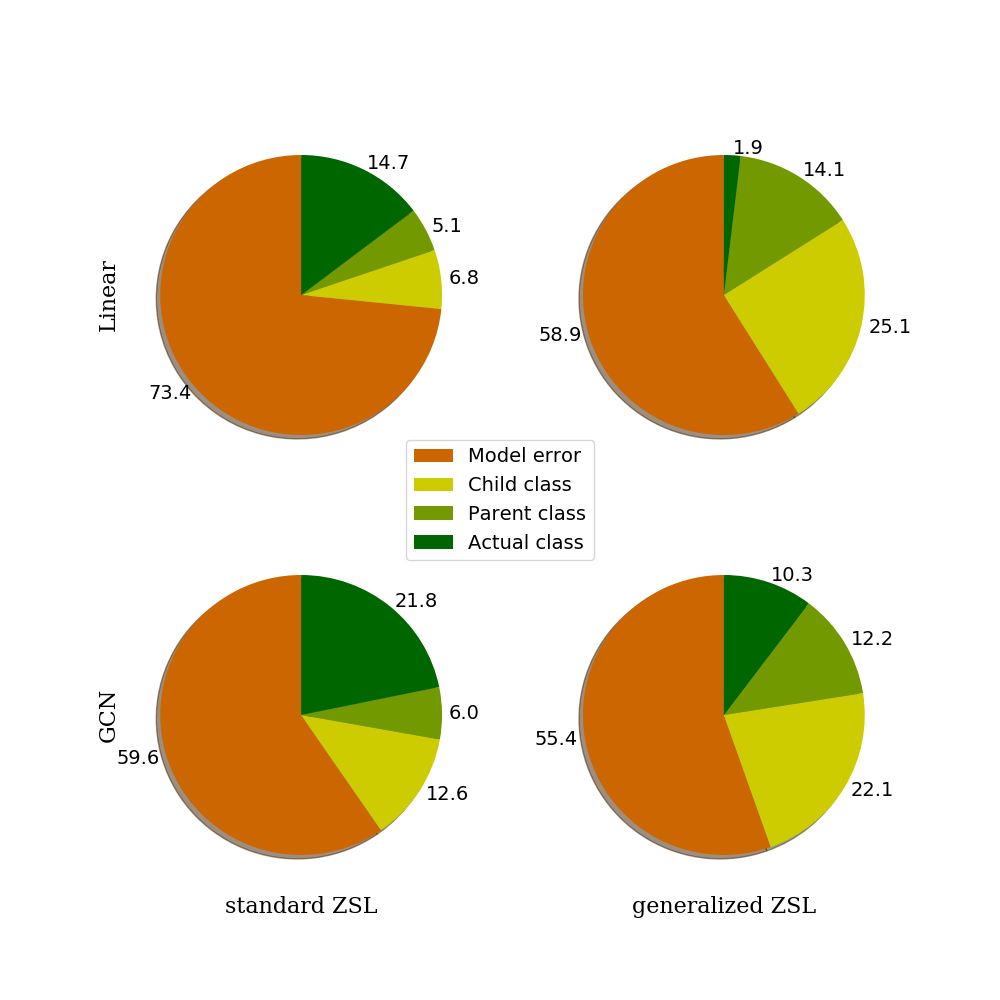}
\caption{
Distribution of the classification outputs of different ZSL models on the \textit{1-hop} test split.
An image $x$ can be either be classified into its actual label $c$, 
the parent class of $c$, one of its children class, or an unrelated class.
Only the latter case constitutes a definitive error.
}
\end{figure}

% Introduce results
We refer to both of the above cases as false negatives.
Figure 2 illustrates the distribution of ZSL classification outputs among these different scenarios on the 1-hop test split.
On the standard ZSL task for instance, the reported accuracy of the GCN model is 21.8\% 
while the actual (semantically correct) accuracy should be somewhere in between 27.8\% and 40.4\%.

% Increase with Generalized
The ratio of false negatives per accuracy increases dramatically in the generalized ZSL setting.
The linear baseline reported accuracy is only 1.9\%, while the actual (semantically correct) accuracy lies between 16.0\% and 41.1\%.
This is due to the fact that ZSL models tend to classify test images into their parent or children training class:
for example, \textit{Cathartid} images tend to be classified as \textit{Vulture}.
Appendix A of the supplementary material presents results on the other standard splits 
on which we show that the ratio of false negative per reported accuracy further increases with with larger test splits.

\subsection{Word embeddings}

% Prelude
In this section, we identify two factors impacting the quality of word embeddings 
and analyse their affect on ZSL accuracy: polysemy and occurrence frequency.
These problems naturally arise in the definition of large scale object categories
so they are inherent problems of ZS recognition of generic objects.
However, we argue that ZSL benchmarks should provide a curated environment with high quality, unambiguous, semantic representations
and that solutions to tackle the special case of polysemous and rare words should 
be separately investigated in the future.

\subsubsection{Occurrence frequency}

% Word embedding scarcity
Word embeddings are learned in an unsupervised manner from the co-occurrence statistics of words in large text corpora.
Common words are learned from plentiful statistics so we expect them to provide more semantically meaningful representations than rare words, 
which are learned from scarce co-occurrence statistics.
We found many Imagenet class labels to be rare words (see Appendix B of the supplementary materials), with as many as 33.7\% of label words appearing less than 50 times in Wikipedia. 
Here, we question whether the few co-occurrence statistics from which such rare word embeddings are learned actually provide any visually discriminative information for ZSL.
%Imagenet situation
%Figure 3 shows the occurrence count distribution of the words labeling Imagenet visual classes in the English Wikipedia corpus. 
%This figure shows a surprisingly high ratio of rare words with as many as 9.7\% of words appearing less than 10 times in Wikipedia. 

% Experiment
To answer this question, we evaluate ZSL models on different test splits of 100 classes:
we split the Imagenet classes into different subsets based on the occurrence frequency of their label word.
We independently evaluate the accuracy of our model on each of these splits and report the ZSL accuracy
with respect to the average occurrence frequency of the visual class labels in Figure 3.

\begin{figure}[h]
\includegraphics[width=0.5\textwidth]{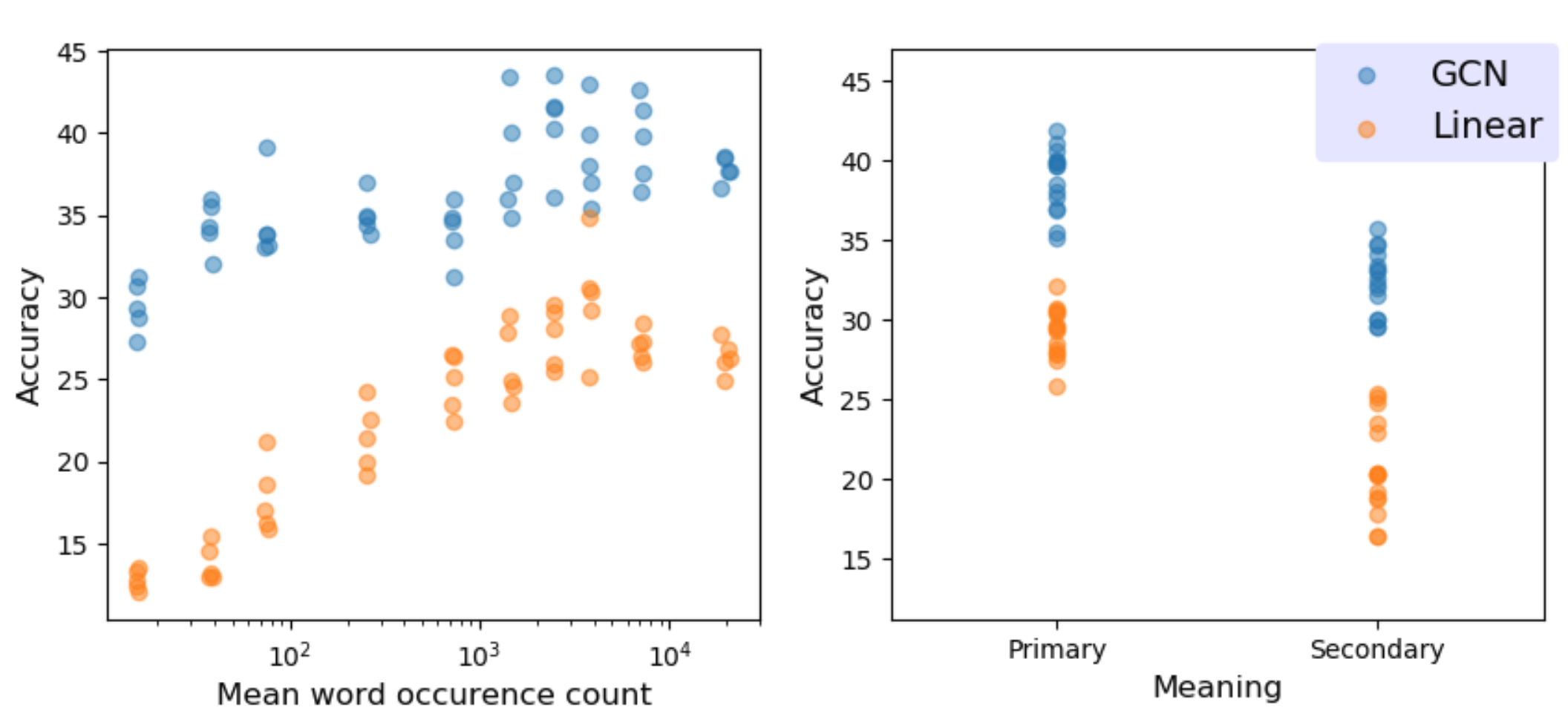}
\caption{
Each dot in these figures represent the top-1 accuracy (y-axis) of a 100 classes test split with respect to the test split characteristics (x-axis):
Left:  Mean occurrence frequency of the test class labels.
Right: test classes of primary meaning, such as cairn (monument), or secondary meaning, such as cairn (dog)
}
\end{figure}

% Describe results
Our results highlight a strong correlation ($r=0.89$) between word frequency and the Linear baseline accuracy 
as test splits made of rare words strikingly under-perform test splits made of more common words,
although accuracy remains well above chance (1\%), even for test sets of very rare words.
Results are more nuanced for the GCN model (correlation coefficient $r=0.74$), which can be explained by the 
fact that GCN uses the Wordnet hierarchy information in addition to word embeddings.

\subsubsection{Polysemy}

% Problem
The English language contains many polysemous words, which makes it difficult to uniquely identify a visual class with a single word.
We found that half of the ImageNet word labels are shared with at least one other Wordnet concept,
and that 38\% of ImageNet classes share at least one word label with other visual classes.
Figure 4 illustrates the example of the word "cairn". 
Two visual classes share the same label "cairn":
One relates to the meaning of cairn as a stone memorial, while the other refers to a dog breed.
This is problematic as both of these visual classes share the same representation in the label space,
so they are essentially defined as the same class although they correspond to two visually very distinct concepts.

% Solution
To deal with polysemy, we assume that all words have one primary meaning, with possibly several secondary meanings.
We consider word embeddings to reflect the semantics of their primary meaning exclusively, and discard visual classes associated with the secondary meanings of their word label.
%Figure 4 shows the t-SNE plot of the word embedding of cairn and vocabulary related to both its meaning.
%This plot suggests that the co-occurrence statistics of the primary meaning of cairn outweights the statistics of its secondary meaning as a dog breed.
To automatically identify the first meaning of visual class labels.
we implement a solution based on both Wordnet and word embeddings statistics detailed in the supplementary material.

\begin{figure}[h]
\centering
\includegraphics[width=0.35\textwidth]{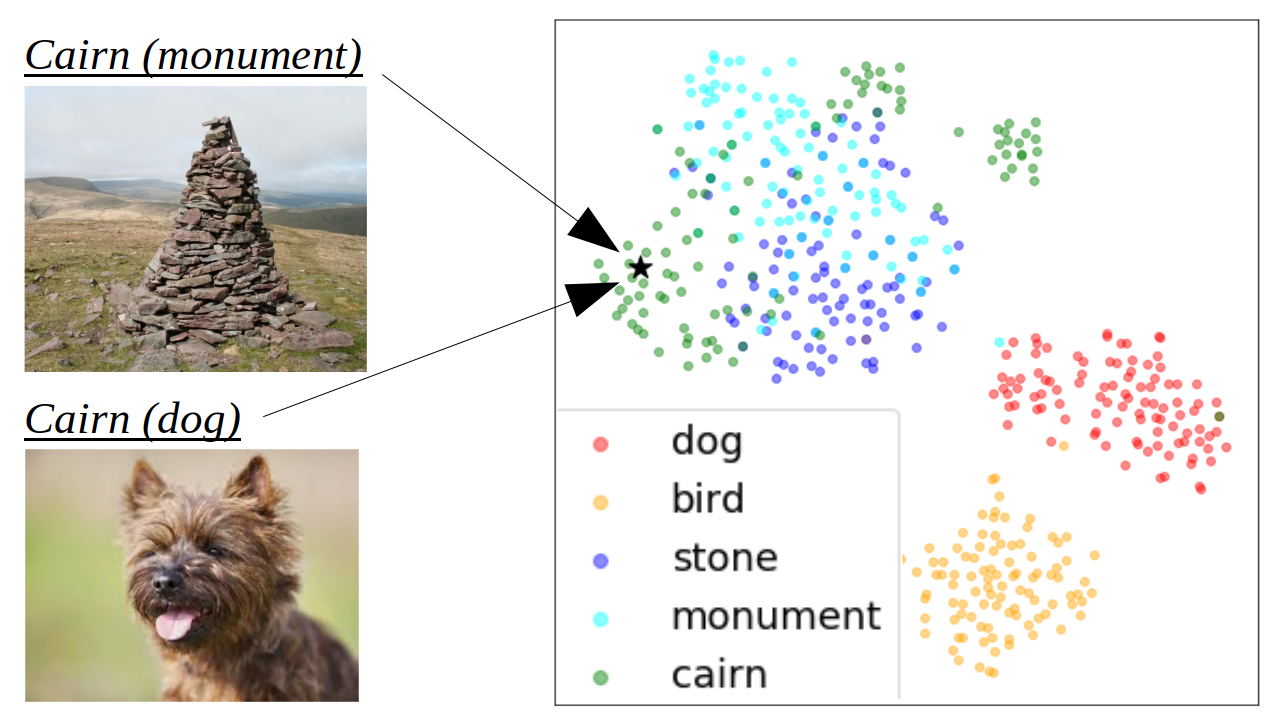}
\caption{
Illustration of polysemous words.
Each color represents the 100 nearest neighbors of a given word.
"Cairn" and its closest neighbors are clustered around the stone and monument related vocabulary, 
far away from dog-related vocabulary so we assign the top visual class as primary meaning of the word cairn.% and the dog class as secondary meaning.
}
\end{figure}

% Experiment
We conduct an experiment to assess both the impact of polysemy on ZSL accuracy and the efficiency of our solution.
As in the previous section, we evaluate our ZSL models on different test splits of 100 classes:
We separately evaluate test classes identified as the primary meaning of their word label and 
test classes corresponding to the secondary meaning of their word label.
Figure 3 reports the accuracy obtained on these different test splits. 
We can see a significant boost in the ZSL accuracy of test classes whose word labels are identified as primary meanings.
In comparison, test splits made exclusively of secondary meanings performed poorly. 
This confirms that polysemy does indeed impact ZSL accuracy,
and suggests that our solution for primary meaning identification allows addressing this problem.

\subsection{Image samples}
% Quality of images
The ILSVRC dataset consists of a high-quality curated subset of the Imagenet dataset.
The current ZSL benchmark uses ILSVRC classes as training classes and classes drawn from the remainder of the Imagenet dataset as test sets,
assuming similar standards of quality from these test classes.
Upon closer inspection, we found these test classes to contain many inconsistencies and ambiguities.
In this section, we detail a solution to automatically filter out ambiguous samples so as 
to only select quality samples for our proposed benchmark.
%A quality benchmark should ensure image samples of high quality.
%We first perform a class-wise selection of images based on the population counts of classes.
%We then use supervised learning to assess the quality of individual images and operate a sample-wise selection.
%We find that the selected high-quality image samples lead to much higher ZSL accuracy, 
%which suggests that much of the reported errors of ZSL models is due to the poor quality of the standard benchmark images.

\subsubsection{Class-wise selection}
% Explanation
Xian \etal \cite{xian2016latent} have first identified a correlation between the sample population of visual classes and their classification accuracy.
They conjecture that small population classes are harder to classify because they correspond to fine-grained visual concepts,
while large population classes correspond to easier, coarse-grained concepts. 
Manual inspection of these classes lead us to a different interpretation:
First, we found no significant correlation between sample population and concept granularity (Appendix C).
For example, fine-grained concepts such as specific species of birds or dogs tend to have high sample populations.
On the other hand, we found many visually ambiguous concepts such as "ringer", "covering" or "chair of state" to have low sample populations.
Such visually ambiguous concepts are harder for crowd-sourced annotators to reach consensus on labeling, resulting in lower population counts.

% Experiment
In Figure 5, we report the ZSL accuracy of our models on different test splits with respect to their average population counts.
This figure shows a clear correlation between the sample population and the accuracy of both models, 
with low accuracy for low sample population classes.
We use the sample population as a rough indicator to quickly filter out ambiguous visual classes and 
only consider classes with sample population superior to 300 images as valid candidate classes in our proposed dataset.

\begin{figure}[h]
\centering
\includegraphics[width=0.45\textwidth]{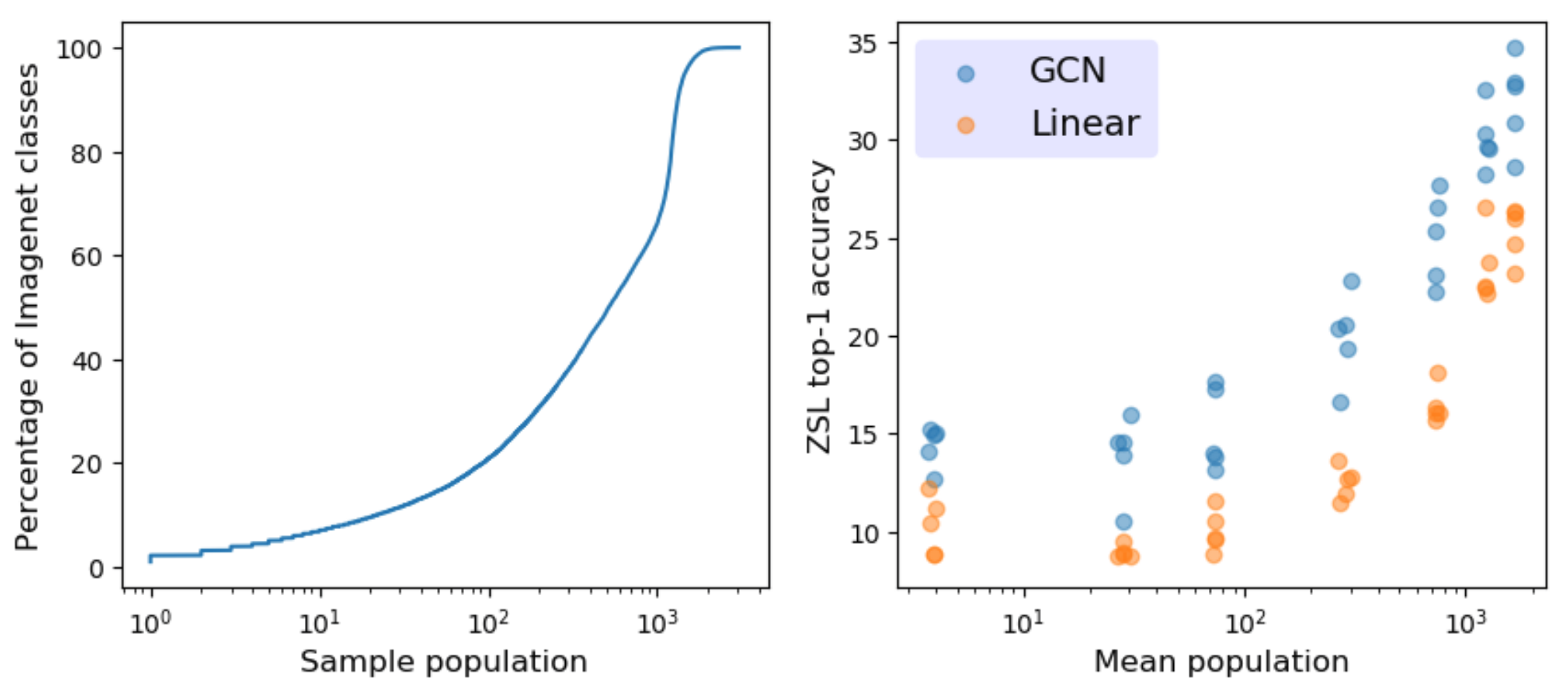}
\caption{
ZSL accuracy with respect to sample population sizes. 
Left: Distribution of Imagenet class population size. 
6.1\% of Imagenet classes have less than 10 samples, 21.1\% have less than 100 samples.
Right: ZSL accuracy of different test splits with respect to their mean sample population size.
}
\end{figure}

\subsubsection{Sample-wise selection} 

% Observation.
Even among the selected classes, we found many inconsistent and ambiguous images to remain (Appendix C), 
so we would like to further filter quality test images sample-wise. % in addition to the class-wise selection.
But what makes a good candidate image for a ZSL benchmark? How can we measure the quality of a sample?
We argue that ZSL benchmarks should only reflect the \textit{zero-shot ability} of models:
ZSL benchmarks should evaluate the accuracy of ZSL models \textit{relatively to the accuracy of standard non-ZSL models}. %In other words, ZSL accuracy should be evaluated relatively to the accuracy of standard, non-ZSL, supervised classification models.
Hence, we define a good ZSL sample as an image unambiguous enough to be correctly classified by standard image classifiers trained in a supervised manner. 

To automatically filter such quality samples, 
we fine-tune and evaluate a standard CNN in a supervised manner on the set of candidate test classes.
We consider consistently miss-classified samples to be too ambiguous for ZSL 
and only select samples that were correctly classified by the CNN
Details of this selection process are presented in Appendix C of the supplementary material. %and additional experiments
%We propose to use supervised learning to automatically filter such quality samples. % from the selected visual classes
%as candidate samples for our ZSL benchmark,
%and we consider the  .
%on a 1000-classes classification task, 
%similarly to the ILSVRC challenge, on random subsets of our candidate classes.

\subsection{Dataset Summary}
% Notice
Figure 6 summarizes the impact of the different factors we analyzed on the top-1 classification error of both our baseline models on the "1-hop" test split.
The error rate of the Linear model on the standard ZSL setting drops from 86\% to 61\% after removing ambiguous images, semantic samples, and structural flaws.
The error rate of the GCN model on the generalized setting drops from 90\% to 47\%.

\begin{figure}[h]
\centering
\includegraphics[width=0.4\textwidth]{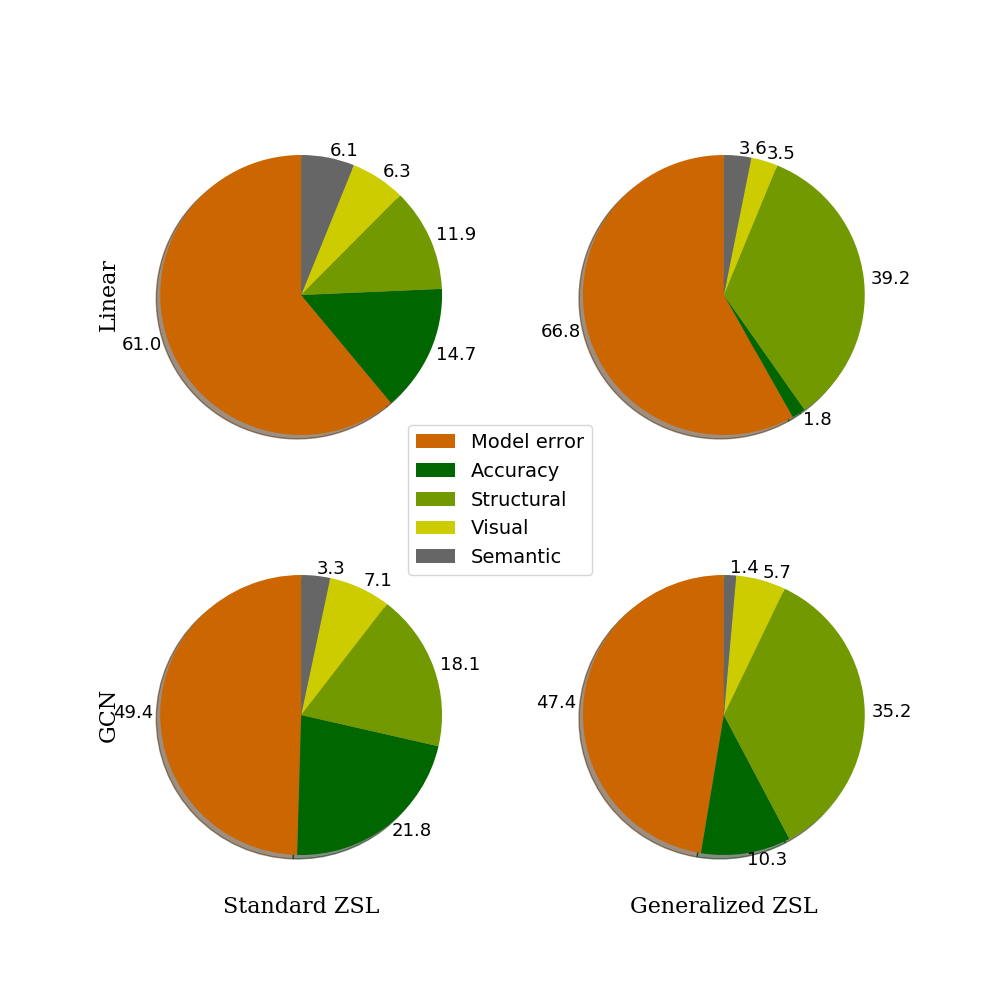}
\caption{
Estimation of the impact of different factors on the reported error of existing models on the 1-hop test split
}
\end{figure}

% Explanation of the results
The GCN model is particularly sensitive to the structural flaws of the standard benchmark, 
but less sensitive to noisy word embeddings than the linear baseline.
This can be easily explained by the fact that GCN models rely on 
the explicit Wordnet hierarchy information as semantic data in addition to word embeddings.
Additional results and details on the methodology of our analysis are given in Appendix D of the supplementary material.

\section{Structural bias}

% Prelude
ZSL models are inspired by the human ability to recognize unknown objects from a mere description, %.
%A commonly given example to illustrate the feasibility of zero-shot learning is as follows:
as it is often illustrated by the following example:
Without having ever seen a zebra, a person would be able to recognize one, 
knowing that zebras look like horses covered in black and white stripes.
This example illustrates the human capacity to \textit{compose} visual features of 
\textit{different} known objects to define and recognize previously unknown object categories.

% Compositionality and Distributed representations
Standard image classifiers encode class labels as \textit{local representations} (one-hot embeddings), 
in which each dimension represents a different visual class, as illustrated in Figure 8. 
As such, no information is shared among classes in the label space:
visual class embeddings are equally distant and orthogonal to each other.
The main idea behind ZSL models is to instead embed visual classes into \textit{distributed representations}:
In label space, visual classes are defined by multiple visual features (horse-ish shape, stripes, colors) shared among classes.
Distributed representations allow to define and recognize unknown classes 
by \textit{composition} of visual features shared with known classes,
in a similar manner as the human ability described above.

% Intro
The embedding of visual classes into distributed feature representations is especially powerful 
since it allows to define a combinatorial number of test classes by composition of a possibly 
small set of features learned from a given set of training classes.
Hence, we argue that the key challenge behind ZSL is to achieve ZS recognition of unknown classes 
by composition of known visual features, following their original inspiration of the human ability, 
and as made possible by distributed feature representations. % based on distributed representation of visual classes.
In this section, we will see that not all ZSL problems require such kind of compositional ability.
On the standard benchmark, we show that a trivial solution based on local 
representations of visual classes outperform existing approaches based on word embeddings.
We show that this trivial solution is made possible by the specific configuration of the standard test splits
and introduce the notion of structural bias to refer to the existence of such trivial solutions in ZSL datasets.

\subsection{Toy example}

% Toy example trivial solution
Figure 7 illustrates a toy ZSL problem in which,
given a training set of \textit{Horse} and \textit{TV monitor} images, 
the goal is to classify images of \textit{Zebra} and \textit{PC laptop}.
Let's consider training an image classifier on the training set and directly applying it to images from the test set.
We can safely assume that most zebra images will be classified as horses, and most laptop samples as TV monitors.
Hence, a trivial solution to this problem consists in defining a one to one mapping between test classes and their closest training class: 
\textit{Horse=Zebra} and \textit{TV monitor=PC laptop}.
This example makes it fairly obvious that not all ZSL problems require the ability to compose visual features to solve. 

\begin{figure}[h]
\includegraphics[width=0.5\textwidth]{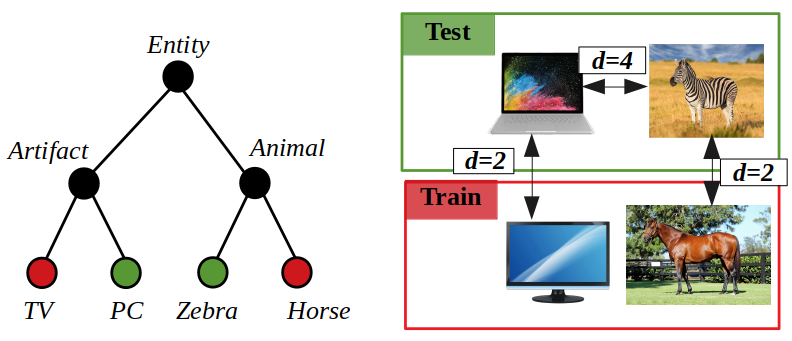}
\caption{
Illustration of the toy example.
Left: Wordnet-like class hierarchy. 
Training classes are shown in red and test class in green.
Right: Illustration of image samples. 
The black captions represent the distance between classes as their shortest path length.
}
\end{figure}
% in the hierarchy.

% Close world and similarity
%Two factors made this trivial solution possible.
%First, 
Classification problems define a close-world assumption:
As all test samples are known to belong to one of the test classes,
classifying an image $x$ into a given test class $c$ means that $x$ is more likely to belong to $c$ than other classes of the test set.
In other words, classification is performed relatively to a negative set of classes \cite{torralba2011unbiased}. 
%Torralba \etal \cite{torralba2011unbiased} discusses the impact of negative set bias in generic object recognition datasets.
%In our toy example, test classes are very dissimilar so their negative set is very distant.
What made this trivial ZSL solution possible is the fact that test classes of our toy example are very similar to one of the training class, relatively to their negative set.
This allowed us to identify a one-to-one mapping by similarity between training and test classes.
We refer to this trivial solution as a similarity-based solution, in opposition to solutions based on the composition of visual features.

\begin{figure}[h]
\centering
\includegraphics[width=0.35\textwidth]{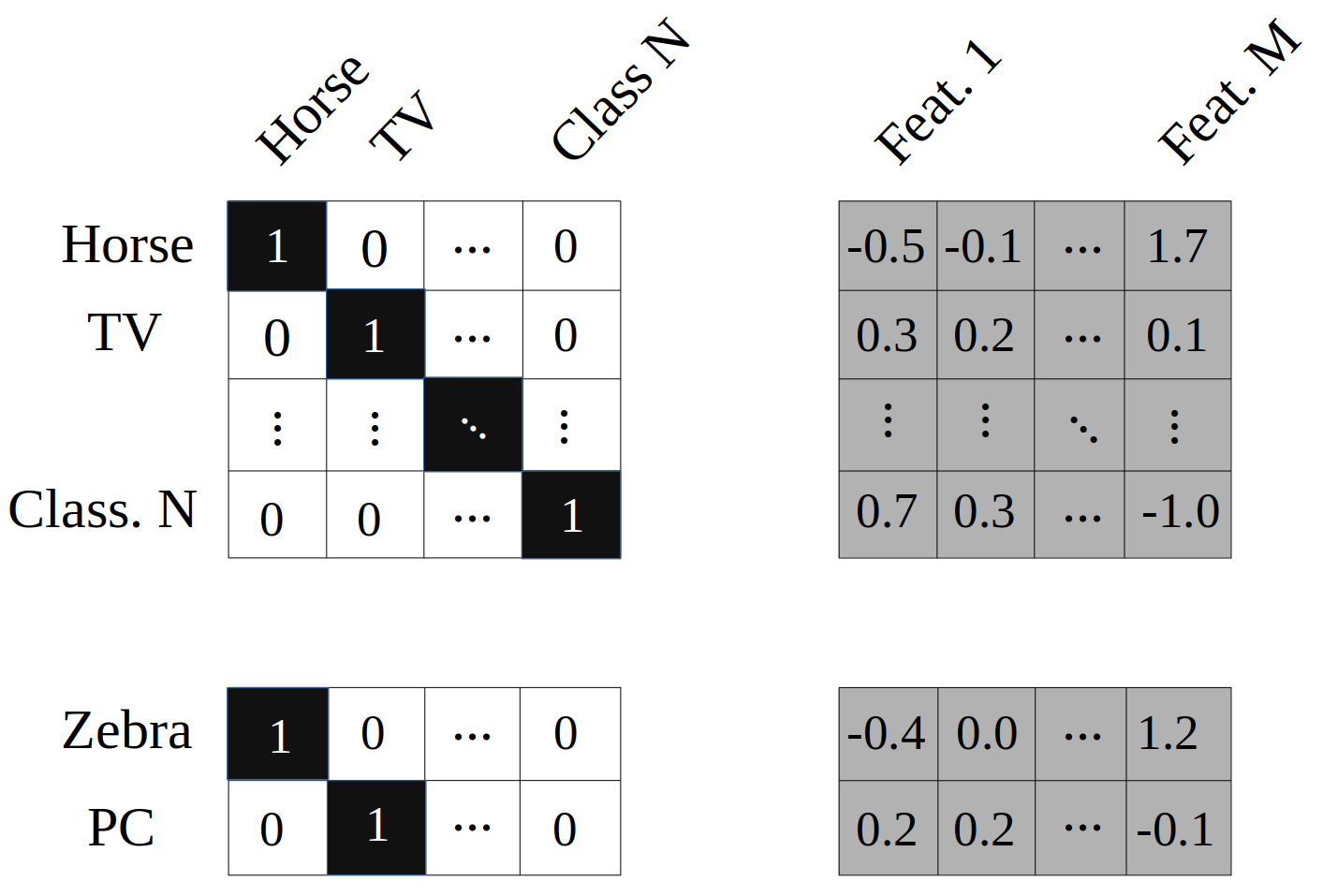}
\caption{
Illustration of local (one-hot, on the left) and distributed (right) representations of visual classes.
The similarity-based solution encodes both training and test classes as local representations.
Composition-based solutions need distributed representations.
%Local representation values are all equal to 0, except for the dimension of their most similar training class.
%No information is shared among training classes as training class embeddings are orthogonal to each other.
%Only one information is shared between training and test classes: a one-to-one similarity mapping.
%Distributed representations can either be interpretable visual features as manually annotated visual attributes or real-values semantic vectors.
%With distributed representations, information is shared among all classes in all dimensions.
}
\end{figure}

% ZSL framing.
%We refer to this trivial solution as a \textit{similarity}-based solution, 
%in opposition to solutions based on \textit{composition} of visual features. 
%The similarity-based solution is trivial to implement within existing ZSL framework:
%we effectively define these two classes as the same.
%Hence, images of Zebra classified as Horse within the training set are classified as Zebras within the test set.
%Horse and Zebra are assigned the exact same embedding in semantic space so that they are essentially defined as the same class.
%These local representations can be used instead of word embeddings in 
% instead of distributed feature representations, one can apply 
%Instead of rich distributed feature representation, 
%one can embed visual classes local representations 
%Figure 9 illustrates the implementation of this solution within the ZSL framework described in Section 3.
%Instead of distributed representations of visual classes, one can assign semantic representations to test classes that only reflect their most similar training class. 
%Classification can be performed using existing ZSL models using these local representations as semantic features instead of distributed representations like word embeddings.

As illustrated in Figure 8, the similarity mapping between test and training classes can be directly embedded in the semantic space using local representations.
The trivial solution consists in assigning to test classes the exact same semantic representation as their most similar training class. 
Consider applying these semantic embeddings within a ZSL framework to our toy problem:
classifying a test image $x$ as a Horse relatively to the negative set of TV within the training set 
becomes strictly equivalent to classifying $x$ as Zebra relatively to its negative set PC within the test set.
Hence, any existing ZSL model using these local embeddings instead of distributed representations like word embeddings $Y$ would converge to the same solution.

\subsection{Standard benchmark}

% Embedding scheme
Besides our toy example, how well would this trivial solution perform on the standard benchmark?
To implement it, we used the Linear baseline model \cite{romera2015embarrassingly} 
with local representations inferred from the Wordnet hierarchy (see Appendix E), 
but any model would essentially converge to a similar solution.
Table 1 compares the accuracy of this trivial solution to state of the art models as reported in \cite{xian2017zero,kampffmeyer2018rethinking}. %compares the accuracy of the trivial similarity-based solution to existing models.
The trivial similarity-based solution outperforms existing ZSL models by a significant margin.
Only GCN-based models \cite{kampffmeyer2018rethinking}, 
which we discuss in the next section, 
seem to outperform our trivial solution.

%We implement the trivial solution using the same baseline linear model \cite{XXX} with local representations (Figure 8) instead of word embeddings as semantic representations $Y$. 
%To compute the local representations, we rely on the Wordnet hierarchy to define one-to-one mappings between test and training classes, as detailed in Appendix 2.
%\begin{subequations} 
%\begin{align}
%&f : C_{te} \rightarrow C_{tr} \\
%&f : c \rightarrow argmin_{c' \in C_{tr}}d(c,c') \\
%&\forall c \in C_{te}, y_c=y_{f(c)}   %\\
% c' &= argmin_{c' \in C_{tr}}d(c,c')
%\end{align}
%\end{subequations} 
%Where $d$ represents the shortest path length between class $c$ and $c'$ within the Wordnet hierarchy.
% Describe results

%We believe that the main reason behind the success of GCN models is that 
%they successfully leverage the structural bias in standard test splits as they use the explicit class hierarchy information, 
%but an in-depth analysis of these models is beyond the scope of this paper.
%It seems likely that GCN-based models perform so well specifically because 
%they make use of the Wordnet hierarchy we used to define the one-to-one mapping 
%of our trivial solution, although this is not 
%However, taking into account the structural flaws of the benchmark as mentioned in section XXX,  
%we realize that our method even outperforms GCN in terms of semantically accurate classification.
%Figure XXX compares the classification output of local representations, GCN and XXX.

\begin{table}[h!]
\centering
\caption{Top-1 accuracy on the standard test splits (top) as reported for linear baselines in \cite{xian2017zero}, (middle) as reported for GCN-based models in \cite{kampffmeyer2018rethinking} and (down) obtained by our trivial solution}
\begin{tabular}{c c c c  }
\hline
model& 1-hop & 2-hops & all \\
\hline
SYNC  \cite{changpinyo2016synthesized}  & 9.26 & 2.29 & 0.96 \\
CONSE \cite{norouzi2013zero}            & 7.63 & 2.18 & 0.95 \\
ESZSL \cite{romera2015embarrassingly}   & 6.35 & 1.51 & 0.62 \\
LATEM \cite{xian2016latent}             & 5.45 & 1.32 & 0.5  \\
DEVISE\cite{frome2013devise}            & 5.25 & 1.29 & 0.49 \\
CMT   \cite{socher2013zero}             & 2.88 & 0.67 & 0.29 \\
\hline
GCNZ  \cite{wang2018zero}               & 19.8  &  4.1  & 1.8 \\
ADGPM \cite{kampffmeyer2018rethinking}  & \textbf{26.6}  &  \textbf{6.3}  & \textbf{3.0} \\
\hline
Trivial                                 & 20.27 & 3.59  & 1.53\\
\hline
\end{tabular}
%\caption*{Comparison of the Trivial solution accuracy to state of the art on the standard benchmark.
%Top: baseline models as reported in \cite{xian2017zero}. Middle: s.o.a GCN-based models as reported in \cite{kampffmeyer2018rethinking}}
\end{table}

%Opening
%Why does such a simple solution outperform a decade of development effort in ZSL?
%In the next section, we will argue that the standard benchmark has a strong structural bias that favors similarity-based ZSL models.

\subsection{Measuring structural bias}
% Definition of ratio
In our toy example, we have hinted at the fact that structural bias emerges 
for test sets in which test classes are relatively similar to training classes,  
while being comparably more dissimilar to each other (to their negative set). 
To confirm this intuition, we define the following structural ratio:

\begin{subequations} 
\begin{align}
r(c) &= \frac{min_{c' \in C_{tr}}d(c, c')}{min_{c' \in C_{te}}d(c, c')} \\
R(C_{te}) &=  \frac{1}{|C_{te}|} \sum_{c \in C_{te}} r(c)
\end{align}
\end{subequations} 

%\begin{equation} 
%r(c) = \frac{min_{c' \in C_{tr}}d(c, c')}{min_{c' \in C_{te}}d(c, c')}
%\end{equation} 
%\begin{equation} 
%R(C_{te}) =  \frac{1}{|C_{te}|} \sum_{c \in C_{te}} r(c)
%\end{equation}

In which $c$ represents a visual class, $C_{te}$ and $C_{tr}$ represent test and training sets respectively, 
and $d$ is a distance reflecting similarity between two classes.
Here, $r(c)$ represents the ratio of the distance between $c$ and its closest training class to the distance between $c$ and its closest test class.
In our experiments, we use the the shortest path length between two classes in the Wordnet hierarchy as a measure of distance $d$,
although different metrics would be interesting to investigate as well.
We compute the structural ratio of a test set $R(C_{te})$ as the mean structural ratio of its individual classes.
% Experiment
Figure 9 shows the top-1 accuracy achieved by baseline models on different test sets with respect to their structural ratio $R$.
As for previous experiments, we report our results on test splits of 100 classes.

\begin{figure}[h!]
\centering
\includegraphics[width=0.35\textwidth]{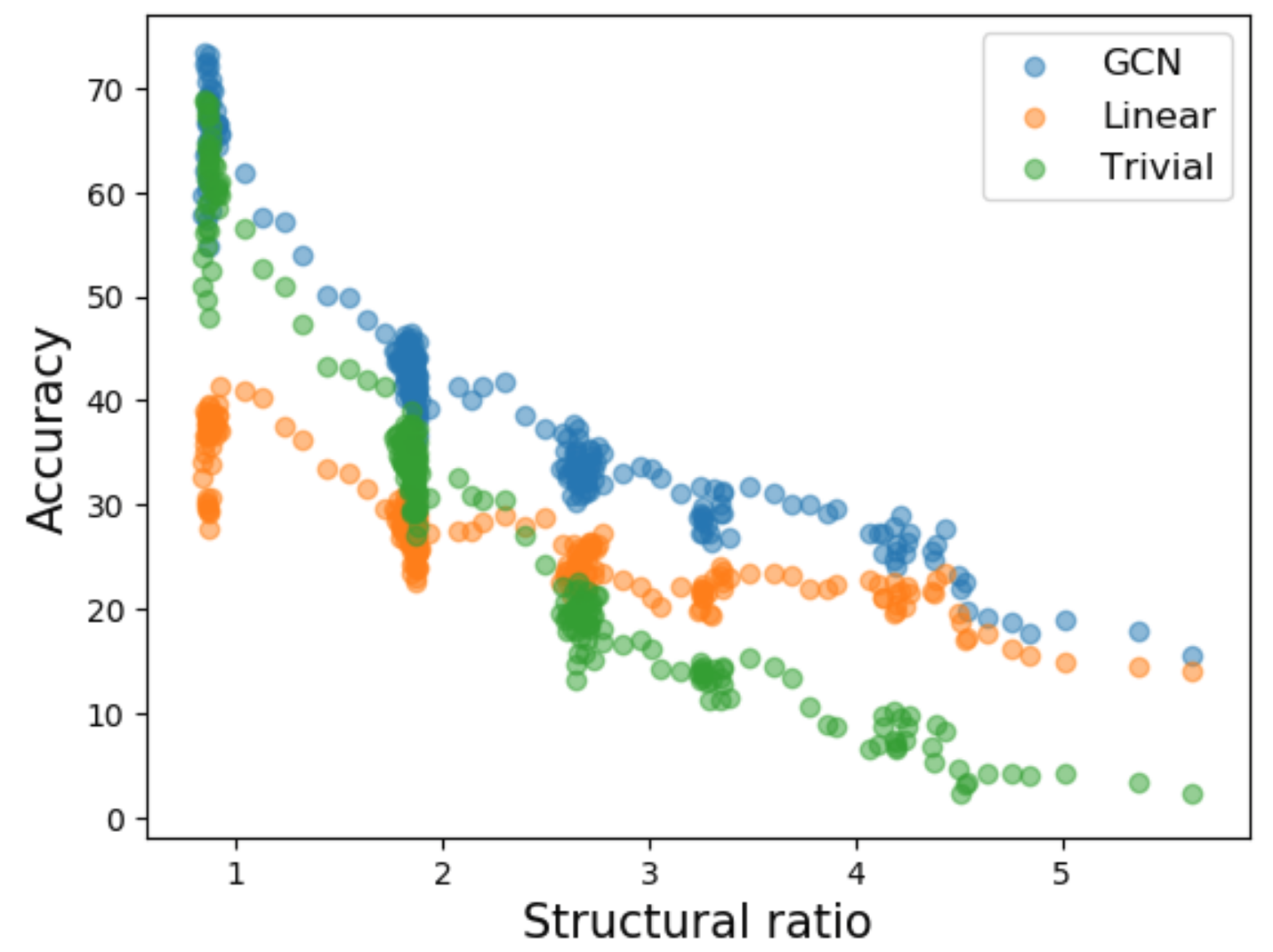}
\caption{
ZSL accuracy on different test sets with respect to their structural ratio $R(C_{te})$.
}
\end{figure}

%Results
On test splits of low structural ratio, the trivial solution performs remarkably well, on par with the state of the art GCN model.
Such test splits are similar to the toy example in which each test class is closely related to a training class while being far away from other test classes in the Wordnet hierarchy.
As an example, the structural ratio of the test split in our toy example is $R(C_{te})=1/2 \times (2/4 + 2/4) = 0.5$, which corresponds to the highest accuracies achieved by the trivial solution.
We say that such test split is structurally biased towards similarity-based trivial solutions.

However, the accuracy of the similarity-based trivial solution decreases sharply with the structural ratio until it reaches near chance accuracy for the highest ratios.
Hence maximizing the structural ratio of test splits seems to be an efficient way to minimize structural bias.
Although their accuracy decrease with larger structural ratios, both GCN and Linear models remain well above chance.
These results suggest that ZSL models based on word embeddings are indeed capable of compositional reasoning.
At the very least, they are able to perform more complex ZSL tasks than the trivial similarity-based solution.
Interestingly, as the trivial solution converges towards chance accuracy, 
the GCN model accuracy seems to converge towards the accuracy of the ZSL baseline.
This suggests that the main reason behind the success of GCN models is that they 
efficiently leverage the Wordnet hierarchy to exploit structural bias.

% Conclusion on the standard and proposed benchmarks
The \textit{1-hop} and \textit{2-hops} test splits of the standard benchmark consist of the set of test classes closest to the training classes within the Wordnet hierarchy.
This leads to test splits of very low structural ratio, similar to our toy example.
For instance, the \textit{1-hop} test split has a structural ratio of 0.55.
It is an example of structural bias even more extreme than 
our toy example as test classes are either children or parent classes of a training class.
In the next section, we propose a new benchmark with maximal structural ratio in order to minimize structural bias.

\section{New Benchmark}

\subsection{Proposed Benchmark}

% Supplementary material
In this section, we briefly detail the semi-automated construction of a new benchmark 
designed to fix the different flaws of the current benchmark highlighted by our analysis.
For space constraints, a number of minor considerations could not be properly presented in this paper.
We detail these additional considerations in Appendix F of the supplementary material. 
Appendix F also provides additional details regarding the different parameters
and the level of automation of each of the construction process.
Appendix G provides details on the code and data we release.
Following Frome \etal \cite{frome2013devise}, we use the ILSVRC dataset as training set, and propose a new test set.
The selection of this new test set proceeds in two steps:

% First step
In a first step, we select a subset of candidate test classes $C' \subset C$ from the remaining 20,845 Imagenet classes based on the statistics of image samples and word labels:
We first filter out semantic samples $Y' \subset Y$ corresponding to rare or polysemous words of secondary meaning (Section 4.2).
We then discard visual classes of low sample population and filter out ambiguous image samples using supervised learning to select $X' \subset X$ (Section 4.3).
The set of candidate test classes is the subset of visual classes $C' \subset C$ for which sufficiently high quality image and semantic samples were selected.

% Second step
In a second step, we define the test split $C_{te} \subset C'$ as a \textit{structurally consistent} set of \textit{minimal structural bias}:
The test set was carefully selected so as to contain no overlap among its own classes nor with the training classes in order to provide a structurally consistent test set for the generalized ZSL setting.
This test set consists of 500 classes of maximal structural ratio $R(C_{te})$ so as to minimize structural bias.

%The training set was selected as a set of 1000 structurally consistent classes (without overlap in the Wordnet hierarchy following Section 4.1) maximally spreaded over the Wordnet hierarchy (Appendix 5).
%The test set consists of 200 classes of maximal strutural ratio $R(C_{te})$ so as to minimize structural bias.
%The test set was carefully selected so as to contain no overlap among its own classes nor with the training classes 
%in order to provide a structurally consistent test set for the generalized ZSL setting.
%so as to only select a subset $X' \subset X$ that are correctly identified by non-ZSL image classifiers
%We filter out classes with rare or polysemic words of second meaning (Section XXX) and classes with low sample population. % under 500 images.
%Following Section 4.3, we also filter out ambiguous image samples so as to only select a subset $X' \subset X$ that are correctly identified by non-ZSL image classifiers.
%Additional classes were manually removed following the additional considerations presented in Appendix XXX of the supplementary materials.

\subsection{Evaluation}

\begin{table}[h]
\centering
\caption{Evaluation on the proposed benchmark. Accuracy in the generalized ZSL setting are reported as harmonic means over training and test accuracy following \cite{xian2017zero}}
\begin{tabular}{ c c c | c c }
\hline
\multirow{2}{*}{Model} & \multicolumn{2}{c}{ZSL} & \multicolumn{2}{c}{G-ZSL} \\
                                                   & @1  & @5      & @1  & @5 \\
\hline
Trivial                                            & 1.2    & 3.9     & 0      & 0 \\
\hline
CONSE            \cite{norouzi2013zero}            & 10.65  & 25.10   & 0.12   &  19.34 \\

DEVISE           \cite{frome2013devise}            & 11.15  & 29.52   & \textbf{7.87}   & 26.10  \\
ESZSL            \cite{romera2015embarrassingly}   & 13.54  & 32.61   & 4.59   & 25.53 \\

\hline
GCN-6            \cite{wang2018zero}               & 9.58   & 27.19   & 4.81   & 23.35 \\
GCN-2            \cite{kampffmeyer2018rethinking}  & 14.09  & 35.12   & 4.96   & \textbf{30.35} \\
ADGPM            \cite{kampffmeyer2018rethinking}  & \textbf{14.10} & \textbf{36.03}  & 4.90 & 29.96 \\
\hline
\end{tabular}
\end{table}

% Results
Table 2 presents the evaluation of a number of baseline models on the newly proposed benchmarks.
A few notable results stand out from this table:
First, different from the standard benchmark, CONSE \cite{norouzi2013zero} performs worse than DEVISE \cite{frome2013devise}.
The relatively high accuracy reported by the CONSE model on the standard benchmark
is most likely due to the fact that word embeddings of test classes are statistically close to
the word embedding of their parent/children test classes so that CONSE 
results more closely fit the trivial similarity-based trivial solution. 
We expect model averaging methods to benefit the most from the structural bias in the standard benchmark.

Second, the impressive improvements reported by GCN-based models over linear baselines are significantly reduced, 
although GCN models still outperform linear baselines. 
This result corroborates the observation, in Section 5, that GCN models 
tend to converge towards the results of linear baseline models for high structural ratio.

\section{Conclusion and Discussion}
% ZSL impact
ZSL has the potential to be of great practical impact for object recognition.
% Dataset
However, as for any computer vision task, the availability of a high quality benchmark is a prerequisite for progress.
In this paper, we have shown major flaws in the standard generic object ZSL benchmark and proposed a new benchmark to address these flaws.
% Goal
More importantly, we introduced the notion of structural bias in ZSL dataset that allows trivial solutions based on simple similarity matching in semantic space.
%We showed that the high degree of structural bias in the current benchmark has favored existing models,
We encourage researchers to evaluate their past and future models on our proposed benchmark.
It seems likely that sound ideas may have been discarded for their poor performance 
relative to baseline models that benefited most from structural bias.
Some of these ideas may be worth revisiting today.

% CCL
Finally, we believe that a deeper discussion on the goals and the definition of ZSL is still very much needed.
There is a risk in developing complex models to address poorly characterized problems:
Mathematical complexity can act as a smokescreen of complexity that obfuscates the real problems and key challenges behind ZSL.
Instead, we believe that practical considerations grounded in common sense are still very much needed at this stage of ZSL research.
The identification of structural bias is a first step towards a sound characterization of ZSL problems.
One practical way to continue this discussion would be to investigate structural bias in other ZSL benchmarks.
%A finer analyse of the interactions between training classes and different test set confi
% Scratch the surface
%whereas the relatively high accuracy of such baselines were mere reflection of the amount of structural bias in the standard benchmark.
%We believe that this work has only scratched the surface of bias in ZSL and the definition or better understanding of ZSL.
%By initiating a discussion on bias in ZSL, we believe that we only scratched the surface.
%In this paper, we have mainly focused on characteristics test sets, with a fixed training set. 
%Future work should also investigate the impact of different 
% Need more practical considerations
%Finally, on a more opinionated note, we do not believe the state of ZSL research to be so advanced as to need ever more complex mathematical solutions to progress.
%While less appealing, we believe that very practical considerations grounded in common sense are still necessary to refine our understanding of the true goals and standards of ZSL.
%As long as a clear and unambiguous, any over-kill in mathematical complexity will only mask the true challenges.

\section*{Aknowledgement}
This work was supported in part by JSPS KAKENHI (Grant No. JP17K00236 
and No. JP17H01995).

{
\small
\bibliographystyle{ieee}
\bibliography{egbib}
}

\newpage 
\onecolumn
\section*{Appendix A. Structural flaws}

Figure 10 reproduces Figure 1 to help the following discussion.
This figure illustrates the configuration of visual classes of the standard test splits within the Wordnet hierarchy.
It should be noted that the \textit{2-hops} test split is a super-set of the \textit{1-hop} split: 
it contains both classes annotated in green and blue.
Similarly, the \textit{all} test split is a super-set of the \textit{2-hops} test split: 
it contains all blue, green and black classes.
In the generalized ZSL setting, training classes (red) are also included in the test set.

\begin{figure}[h!]
\centering
\includegraphics[width=0.65\textwidth]{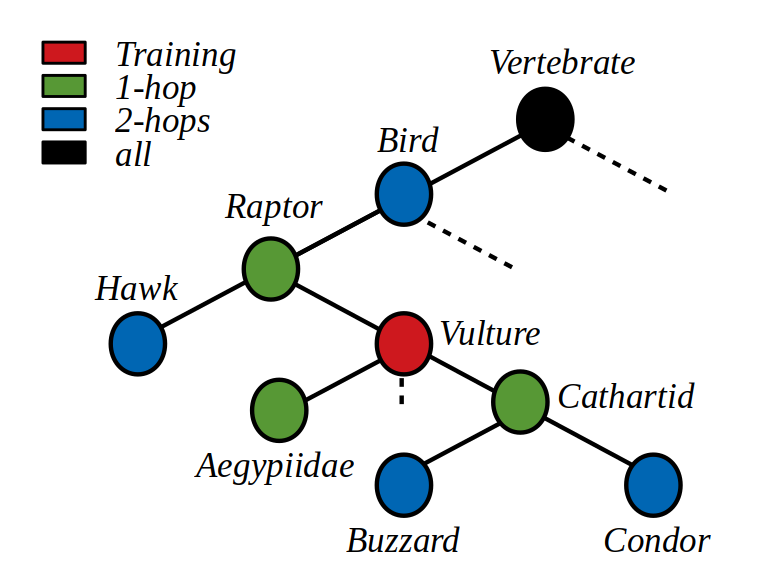}
\caption{
Illustration of the standard test splits configuration.
}
\end{figure}

Figure 11 and 12 illustrate the distribution of ZSL classification outputs on the \textit{2-hops} and \textit{all} test splits respectively.
On the \textit{2-hops} standard ZSL test set, 3.6\% of test images were correctly classified by the Linear baseline model.
This ratio corresponds to the percentage of images of \textit{Raptor} correctly classified as \textit{Raptor}, 
\textit{Buzzard} images classified as \textit{Buzzard}, etc.
We refer to such classification outputs as True Positive (TP).
These correspond to the accuracy reported by previous works on the standard benchmark.
2.3\% of test images were classified as one of their parent class:
These correspond to images of \textit{Buzzard} or \textit{Hawk} classified as \textit{Raptor} or \textit{Bird} for example.
These classification outputs are considered as errors by the current benchmark, 
while they are semantically correct: a \textit{Hawk} is just a specific kind of \textit{Bird}.
3.7\% of test images were classified as one of their children class:
images of \textit{Raptor} or \textit{Bird} classified as \textit{Buzzard} or \textit{Hawk}.
Such classification outputs are considered as errors by the current benchmark, 
whereas they may be either semantically correct or incorrect depending on the specific kind of bird in the image.
We refer to both of these classification scenarios as False Negative (FN).
On the other hand, an image of \textit{Buzzard} classified as \textit{Aegypiidae} is an actual classification error:
\textit{Buzzard} and \textit{Aegypiidae} are two distinct, mutually exclusive concepts.
We refer to such classification errors as True Negatives (TN).

\begin{figure}[h!]
\centering
\includegraphics[width=0.65\textwidth]{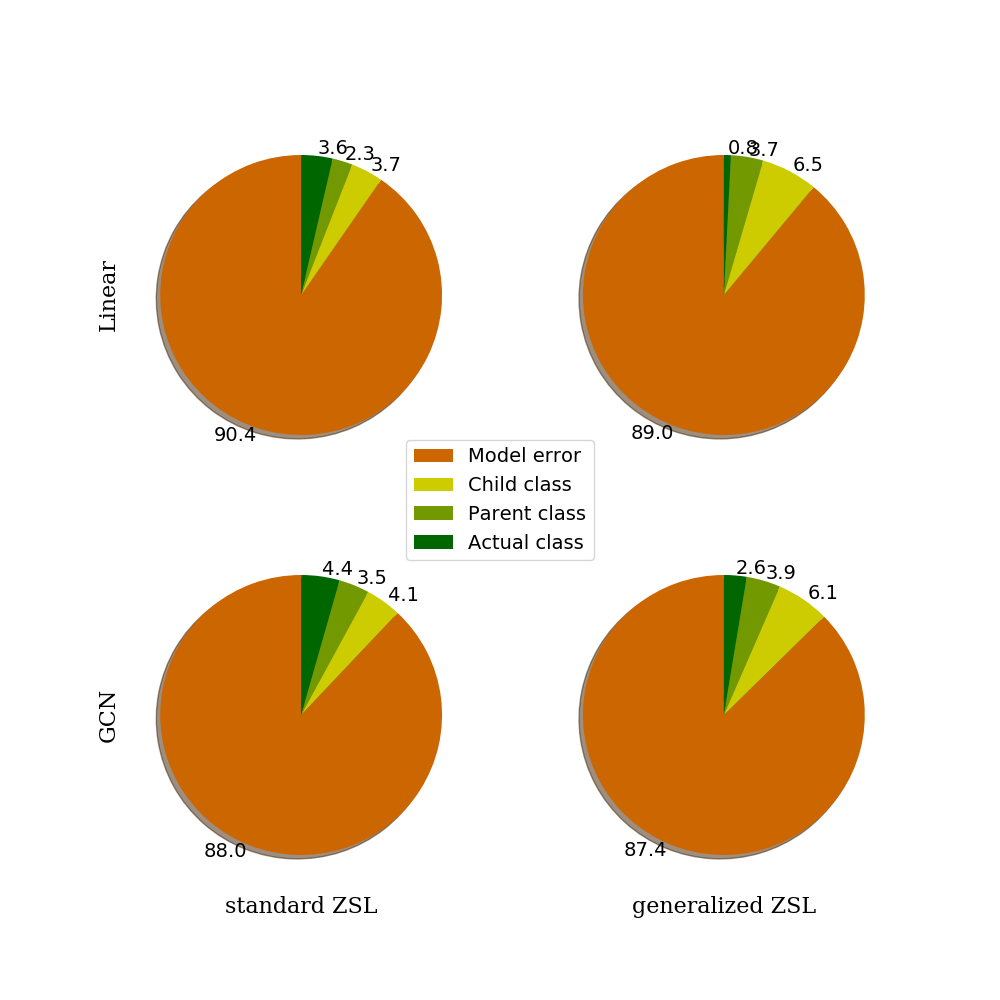}
\caption{
Distribution of classification outputs on the \textit{2-hops} test split.
}
\end{figure}

\begin{figure}[h!]
\centering
\includegraphics[width=0.65\textwidth]{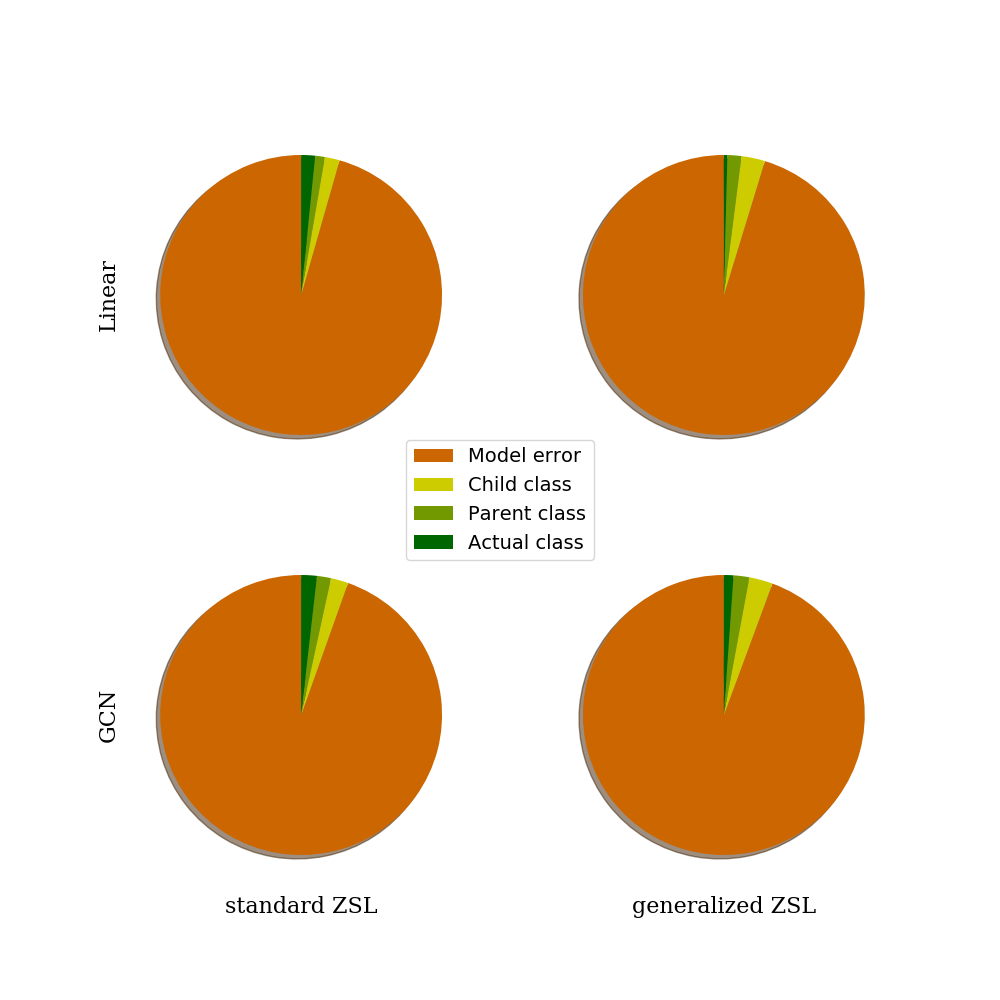}
\caption{
Distribution of classification outputs on the \textit{all} test split.
}
\end{figure}

Table 3 summarizes the ratio of false negative per true positive on each of the standard test split:
$ratio = FN / TP$.
This table shows two interesting trends: 
First, as noted in the original paper, the ratio is much higher in the Generalized ZSL setting.
This is due to the fact that ZSL models tend to classify test images as their parent or children training class.
Second, in the standard ZSL setting, the ratio tends to increase with larger test sets:
the GCN model ratios are 2.3, 3.8 and 4.1 on the \textit{1-hop}, \textit{2-hops} and \textit{all} test splits respectively.
We believe this is due to larger overlaps within the Wordnet hierarchy:
In the \textit{1-hop} test set, the only FN classes for \textit{Cathartid} images is \textit{Raptor}.
In the \textit{2-hops} test set, \textit{Buzzard}, \textit{Condor}, 
\textit{Raptor} and \textit{Bird} are all FN classification outputs for \textit{Cathartid} images.
This trend, however, does not hold for the Linear model in the Generalized ZSL setting.

\begin{table*}[h!]
\centering
\caption{Ratio of false negatives (FN) per true positives (TP).}
\begin{tabular}{ c c | c c c | c c c | c c c }
\multicolumn{2}{c}{}&\multicolumn{3}{c}{1-hop}&\multicolumn{3}{c}{2-hops}&\multicolumn{3}{c}{all} \\
\hline
Model & Task                     & TP     & FN    & ratio            & TP    & FN.   & ratio          & TP   & FN    & ratio\\
\hline        
\multirow{2}{*}{Linear} & ZSL    & 14.7   & 10.2  & \textbf{0.7}     & 3.6   & 6.0   & \textbf{1.7}   & 1.6  & 2.8   & \textbf{1.7}  \\
                        & GZSL   & 1.9    & 39.2  & \textbf{20.6}    & 0.8   & 10.23 & \textbf{12.7}  & 0.4  & 4.27  & \textbf{10.7} \\
\hline
\multirow{2}{*}{GCN}    & ZSL    & 21.8   & 18.6  & \textbf{0.8}     & 4.4   & 7.6   & \textbf{1.7}   & 1.8  & 3.6   & \textbf{2.0}  \\
                        & GZSL   & 10.3   & 34.2  & \textbf{2.3}     & 2.6   & 10.0  & \textbf{3.8}   & 1.1  & 4.5   & \textbf{4.1}  \\
\hline
\end{tabular}
\end{table*}

\section*{Appendix B. Word embeddings}
\subsection*{Occurrence frequency}
We used the full English Wikipedia corpus to estimate the occurrence frequency of words:
we scanned the Wikipedia corpus to count the occurrence of each visual class labels  
(\textit{Hawk}, \textit{Raptor} or \textit{Aegypiidae}, etc.).
We use these occurrence counts as a measure to identify rare and common words.
Figure 13 represents the cumulative distribution of visual class label occurrence counts.

\begin{figure}[h!]
\centering
\includegraphics[width=0.65\textwidth]{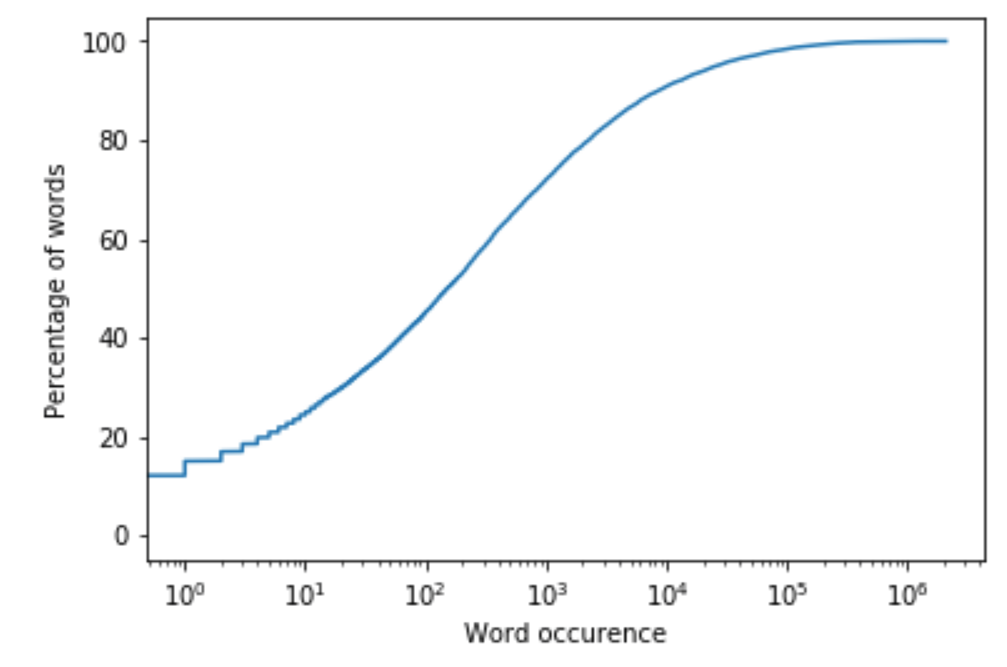}
\caption{
Word occurrence cumulative distribution. The x axis is in logarithmic scale.
}
\end{figure}

As shown in this figure, 24\% of Imagenet class labels occur less than 10 times in the full Wikipedia corpus.
45\% of Imagenet class labels occur less than 100 times.
We found that fine-grain animal species, in particular, exhibit rare word labels (see Figure 10).
We expect the word embedding of such classes to provide noisy semantic representations, 
which has been confirmed by the experiments presented in the original paper.

\subsection*{Polysemy}

\begin{figure}[h!]
\centering
\includegraphics[width=0.65\textwidth]{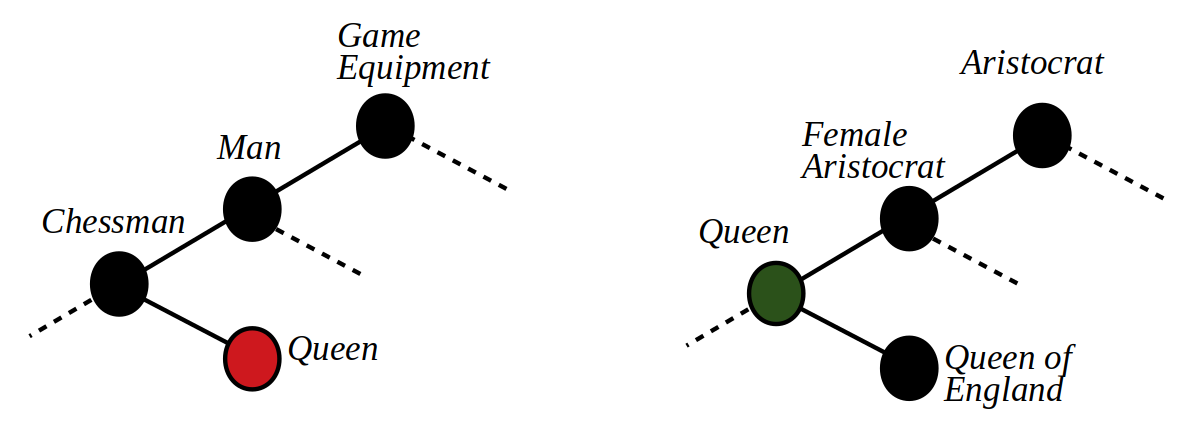}
\caption{
Illustration of two Wordnet concepts sharing the same label Queen.
}
\end{figure}

Figure 18 illustrates several polysemous visual classes of the Imagenet dataset.
To deal with polysemy, we want to assign a unique visual class to polysemous words.
To do so, we define a similarity score $s(w, c)$ between words $w$ and their visual classes $c$.
Given a polysemous word $w$, we assign $w$ to its visual class $c$ of highest similarity score:

\begin{subequations} 
\begin{align}
& s :  W \times C \rightarrow \mathbb{R} \\
& c^* = argmax_{c \in C}s(w, c) 
\end{align}
\end{subequations} 

As a similarity score, we use the cosine similarity between 
word embeddings and the average word embedding of visual class parent and children concepts.
Consider the example of the word $Queen$ illustrated in Figure 14.
There are 9 visual classes associated with the word  $Queen$ in the Imagenet dataset.
For brievity, we only consider two of the $Queen$ visual classes: 
one as an $Aristocrat$, and one as a $chess piece$
The similarity score between $Queen$ and its $Aristocrat$ visual class is given by:

\begin{subequations} 
\begin{align}
& s(c,w) = cos(w_{Queen}, \times(w_{Aristocrat} + w_{Female} + w_{England})/3) \\
& s(c,w) = 0.23
\end{align}
\end{subequations} 

The similarity score between $Queen$ and its $Chess$ visual class is given by:

\begin{subequations} 
\begin{align}
& s(c,w) = cos(w_{Queen}, w_{Chessman}) \\
& s(c,w) = -0.04
\end{align}
\end{subequations} 

So we assign the word $Queen$ to the visual class of highest similarity score: 
The one corresponding to the $Aristocrat$ meaning.

\section*{Appendix C. Visual samples}

\subsection*{Class-wise selection}

Xian \textit{et al.} \cite{xian2017zero} have proposed different test splits based on visual class sample populations.
They conjecture that small population classes correspond to fine-grained visual concepts,
while large population classes correspond to coarse-grained concepts.
Manually inspecting each of these visual classes,
we found many fine-grain concepts to have large image sample populations while many coarse grain concepts have small sample populations.
As a measure of the "granularity" of visual classes, we propose to use their distance to the root node within the Wordnet hierarchy.
Fine-grain classes are lower in the Wordnet hierarchy, hence further away from the root node than coarse-grain classes.

\begin{figure}[h!]
\centering
\includegraphics[width=0.65\textwidth]{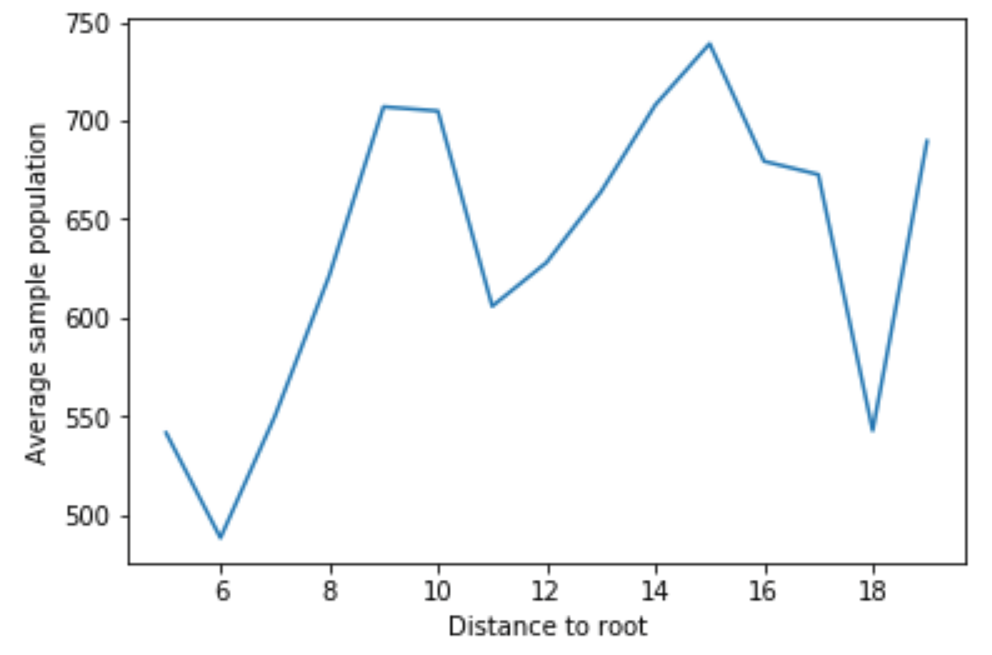}
\caption{
Average sample population per visual class with respect to their "granularity".
}
\end{figure}

Figure 15 shows the average sample population of visual classes with respect to their distance to the root node in the Wordnet hierarchy.
Visual classes within 6 hops of the root node have an average sample population of 490 images.
Visual classes within 10 hops of the root node have an average sample population of 700 images.
This figure illustrates no clear correlation between visual class granularity and their sample population.
In contrast, we found that many low sample population classes instead correspond to visually ambiguous concepts, as illustrated in Figure 19.
Hence, we remove low sample population classes from our proposed benchmark to avoid visually ambiguous concepts.

\subsection*{Sample-wise selection process}

We define high-quality image samples as images that can be correctly classified by a supervised model on a non-ZSL classification task.
We propose a simple procedure to select such image samples.
Given a set of labeled samples $X=\{(x,c)\}$, 
our procedure returns a subset $X' \subset X$ of high-quality images.
This selection process is formalized in Algorithm 1, and proceeds as follows:

First, we randomly sample subsets of 1000 visual classes $C' \subset C$ from the full Imagenet dataset.
Classes are sampled so as to contain no overlap in the Wordnet hierarchy: 
random splits $C'$ do not contain both parent and their children classes.

Second, we randomly sample 250 images per class as training samples, 
and use the remaining images as test samples.
We fine-tune the last layer of a pretrained Resent-50 on the set of training samples, 
and evaluate the classification output of the model on the test samples.

We consider correctly classified image samples as high-quality test samples 
for our benchmark and discard the incorrectly classified images.
We repeat this operation until all samples $x \in X$ have been evaluated.
The output $X'$ of this procedure is a subset of high-quality image samples 
that were correctly classified by the model.

\begin{algorithm}[h]

 \textbf{Input:} \\
  Imagenet Dataset: $X=\{(x,c) \in \mathbb{R}^{3 \times h \times w} \times C\}$\\
  ILSVRC-pretrained ResNet: $BaseModel: \mathbb{R}^{3 \times h \times w} \rightarrow C$ \\
 \textbf{Output:}\\
  High-quality Imagenet subset: $X' \subset X$ \\
 \textbf{Init:} \\
 Initialize an empty error set $Err=\emptyset$ and accurate set: $Acc=\emptyset$\\
 \While{$Err\cup Acc \neq X$}{
  $C'= SampleClass(C, 1000)$\\
  $X_{C'} = \{(x,c) | c \in C'\}$\\
  $X_{train},X_{test} = SampleSplit(X_{C'}, 250)$\\
  $Model = FineTune(BaseModel, X_{train})$ \\
  \For{$(x,c) \in X_{test}$}{
    \eIf{$Model(x)==c$}{
       $Acc= Acc\cup \{(x,c)\}$
    }{
       $Err= Err\cup \{(x,c)\}$
    }
  }

 }
 $X'=Acc$\\
 \textbf{end}

 \caption{Sample-wise selection procedure. 
$SampleSplit(C, n)$ is a sampling procedure that returns a subset $C'$ of n non-overlapping classes (i.e.; no children classes and their parents are contained in $C'$) from the class set $C$.
$SampleSplit(X, n)$ is a sampling procedure that returns a training set $X_{train}$ of n training samples for each class in $X$, and the remaining samples as a test set $X_{test}$.
$FineTune(M, X)$ is a procedure that fine-tunes a model $M$ on the input training set $X$.
}
\end{algorithm}

\section*{Appendix D. Standard benchmark summary}

Figure 15 of the main paper summarizes the impact of visual, semantic and structural flaws
on the \textit{top-1} accuracy of the \textit{1-hop} test split.

In these plots, the accuracy score (in green) corresponds to the model accuracy as reported by the standard benchmark.
The model error (in orange), represents the classification errors after removing ambiguous images, semantic samples, and structural flaws.
For example, the error rate of the GCN model on the generalized setting drops from 90\% to 47\%. 
In order to estimate the impact of all three individual factors individually,
we ran a set of $2^3=8$ experiments with all possible configurations: 
with or without considering visual sample quality, semantic sample quality, and structural flaws.
The estimated impact reported for each factor corresponds to the mean improvement in classification accuracy brought by this specific factor within all the other factors configuration.
Figure 16 and 17 of this supplementary material report similar analysis on the \textit{top-1} accuracy of the \textit{2-hops} and \textit{all} test splits respectively.

\begin{figure}[h!]
\centering
\includegraphics[width=0.68\textwidth]{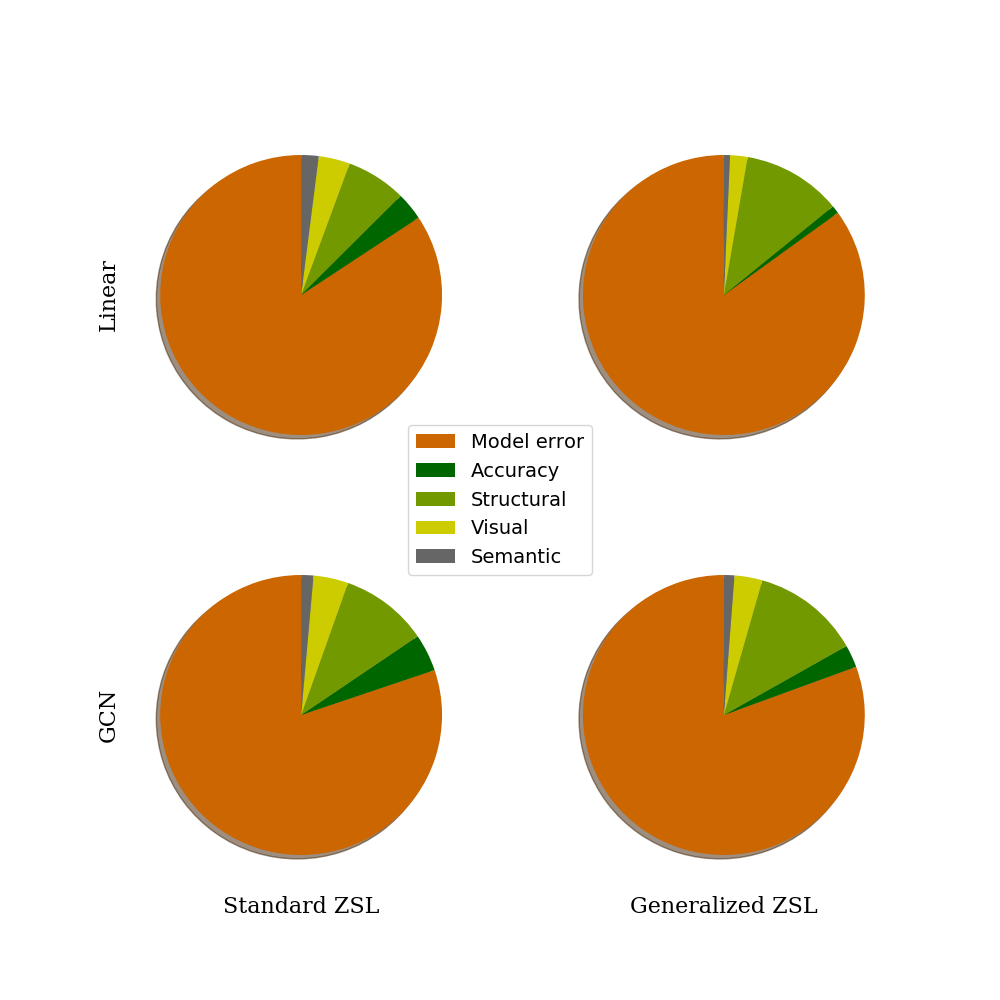}
\caption{
Estimation of the impact of different factors on the
reported error of existing models on the \textit{2-hops} test split.
}
\end{figure}

\begin{figure}[h!]
\centering
\includegraphics[width=0.68\textwidth]{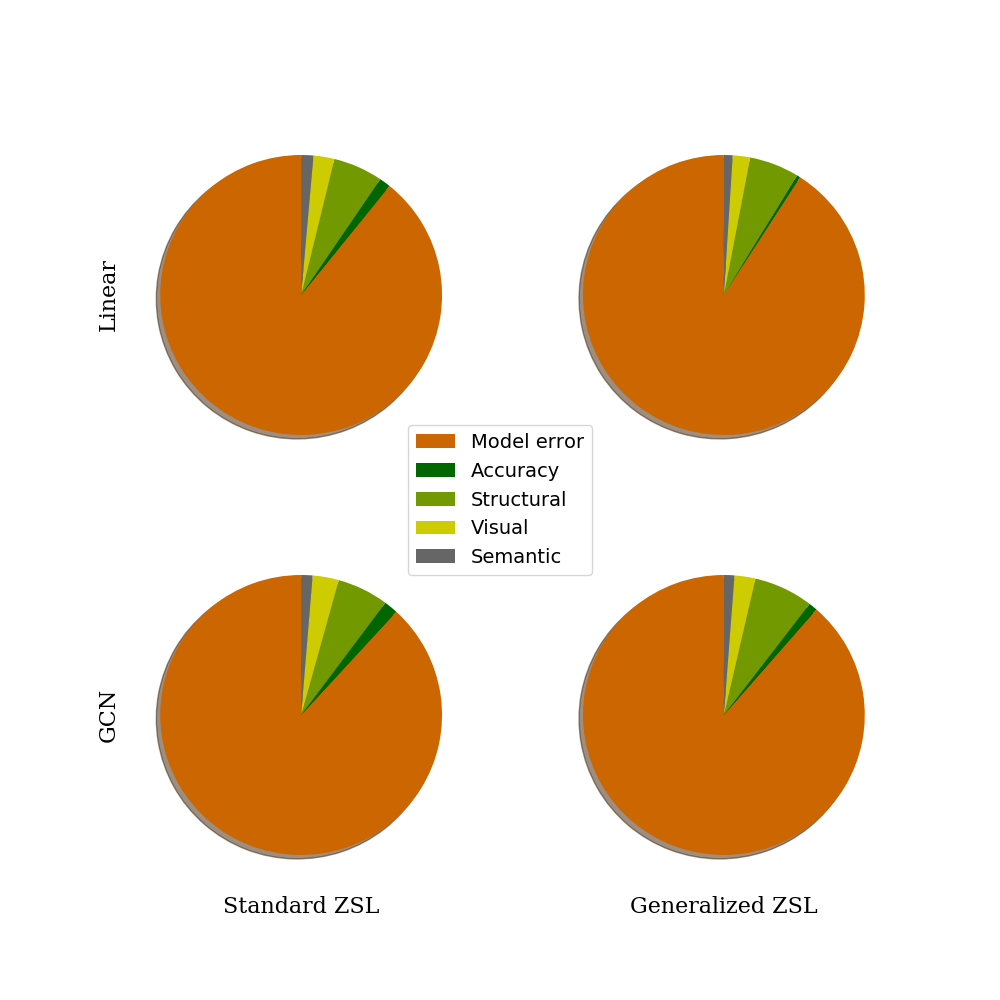}
\caption{
Estimation of the impact of different factors on the
reported error of existing models on the \textit{all} test split
}
\end{figure}

\section*{Appendix E. Trivial solution}

To apply the trivial solution of the toy example to the standard benchmark, we need a similarity mapping $f$ between training and test classes.
To define such mapping, we used the shortest path length between nodes of the Wordnet hierarchy as a measure of distance $d$.
We assign to test classes the semantic embedding of their closest training class, as formalized in equations (4.): 

\begin{subequations} 
\begin{align}
&f : C_{te} \rightarrow C_{tr} \\
&f : c \rightarrow argmin_{c' \in C_{tr}}d(c,c') \\
&y_c=y_{f(c)} + e , \forall c \in C_{te}
\end{align}
\end{subequations} 

However, this procedure leads to many test classes sharing the exact same semantic representations.
Consider the example of \textit{Cathartid} and \textit{Aegypiidae} classes in Figure 10.
Both classes are closest to the \textit{Vulture} training classes so they share the same semantic vector $y_{Voluture}$
This leads to undefined behaviors in the classification process.
To differentiate between such classes, we add a small Gaussian noise $e$ to the semantic embeddings of test classes,
following equation (4c).

The trivial solution can be implemented by any existing ZSL model using these semantic embeddings.
The results reported in the original paper were computed using the Linear baseline.

\section*{Appendix F. Dataset construction}
\subsection*{Additional considerations}

A number of additional factors were taken into consideration in the construction of our proposed benchmark.
For space constraints, we could not include these considerations in the original paper, so we briefly present them in this Appendix.

\textbf{Sample population:}
The number of images per test class in the standard benchmark's test splits is very uneven.
Some test classes have as little as one sample image, while some classes have thousands of images.
This leads to highly biased evaluations as test classes of high sample population 
have a larger impact on the reported classification accuracy.
We select 100 quality samples for each test class to ensure an evenly distributed test set.

\textbf{Mutual exclusion:} 
To prevent false negative classification outputs, test classes should be mutually exclusive.
The hierarchical structure of Wordnet allows us to automatically create test splits 
that do not include both parent and test classes,
so we can automatically remove such mutually non-exclusive classes from the test sets.
However, this is not sufficient to guarantee the mutual exclusivity of test classes.
For example, the Imagenet dataset includes classes such as $Man$, $Woman$, $White$ $Person$, or $Engineer$.
We do not want to include such kinds of classes in our benchmark because classifying an 
image of $Woman$ as $White$ $Person$ or $Engineer$ would result in false negative outputs.
These classes, although not directly related to each other in the Wordnet hierarchy, are not mutually exclusive.
The Wordnet hierarchy does not provide the logical constructs to automatically detect such instances,
so we manually inspect the set of candidate test classes and remove them from the test set. 

\textbf{Scale considerations:}
We favor images of generic objects captured at the scale of human perceptions:
we remove classes of images taken at microscopic scale (biological cells, bacteria, etc.), 
or classes of images at astronomical scales (supernova).

\textbf{Shape considerations:}
We favor objects that can be recognized by their characteristic shape 
and remove classes that require reading comprehension to identify.
For example, we remove a number of medicines, such as $Vitamin D$ or branded contents like $Pepsi$ $Cola$.
Figure 20 illustrates a few such classes.

\subsection*{Dataset construction Summary}

Table 4 summarizes the different steps of the creation of our benchmark.
It details the level of automation, the different parameters involved in each step,
as well as the approximate ratio of visual classes selected within each of these steps.

\begin{table*}[h!]
\centering
\caption{Summary of the benchmark construction steps}
\begin{tabular}{ c c c c c }
                            &  Step                &     Automation     & Parameters                   &     Filter ratio \\
\hline
\multirow{2}{*}{Semantic}   &  Frequency           &     Auto           & $f>500$                      &     82\%    \\
                            &  Polysemy            &     Auto           &  -                           &     91\%    \\
\hline
\multirow{4}{*}{Visual}     &  Class-wise          &     Auto           & $n>300$                      &     63\%    \\
                            &  Sample-wise         &     Auto           & $n_C=1000$, $n_{tr}=250$     &     100\%   \\
                            &  Shape               &     Manual         &  -                           &     95-99\% \\
                            &  Scale               &     Manual         &  -                           &     99\%    \\
\hline
\multirow{2}{*}{Structural} &  Hierarchy           &     Auto           &  -                           &     82\%    \\
                            &  Mutual Exclusivity  &     Manual         &  -                           &     95-99\%    \\
\hline
\end{tabular}
\end{table*}

The majority of the visual classes filtered out from our benchmark were automatically discarded based on their weak semantic features, 
low sample population or structural constraints to avoid both parents and children classes be included in the test set.
Only the semantic and visual sample selection steps are parameterized.
We select word labels occurring at least 500 times within the Wikipedia corpus to avoid rare words.
We only select visual classes with a sample population superior to 300 images.

\section*{Appendix G. Code \& Data}

The full Imagenet dataset, as considered in the \textit{all} test split consists of over 13 million images, 
which is very time-consuming to download and process.
In contrast, small-scale benchmarks like AwA, CUB or SUN
come with off-the-shelf semantic and visual features.
Furthermore, they are orders of magnitude smaller than the Imagenet dataset which makes it much 
easier for researchers to evaluate their models on.
As a result, many recent works on ZSL have only reported the evaluation of their models on small-scale benchmarks,
instead of the standard Imagenet benchmark.

To encourage researchers working on ZSL to evaluate their model on our proposed benchmark,
we release pretrained semantic and visual features\footnote{Download instructions are available at https://github.com/TristHas/GOZ}.
The dataset is small enough to fit in the memory of most modern computer hardware so it allows for fast prototyping and evaluation.
To work on the original raw images, we provide the URL of test images with a Python script for download.

In addition to this data, we also provide code for visual class selection and fast manipulation of the Wordnet hierarchy.
This should allow researchers interested in the investigation of different factors impacting ZSL accuracy to quickly build different test splits.

\begin{figure}[h!]
\centering
\includegraphics[width=0.95\textwidth]{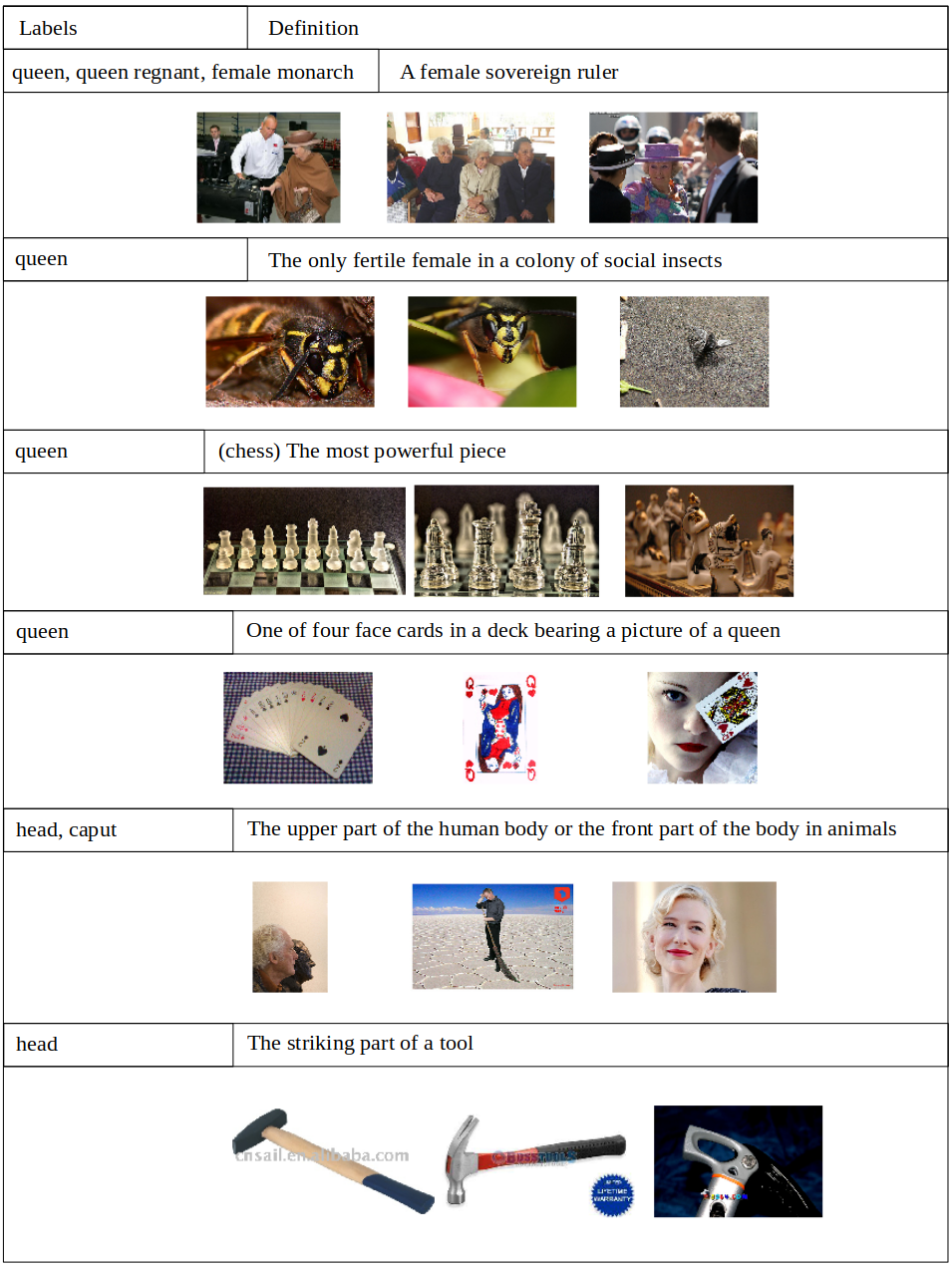}
\caption{
Examples of polysemous classes
}
\end{figure}

\begin{figure}[h!]
\centering
\includegraphics[width=0.95\textwidth]{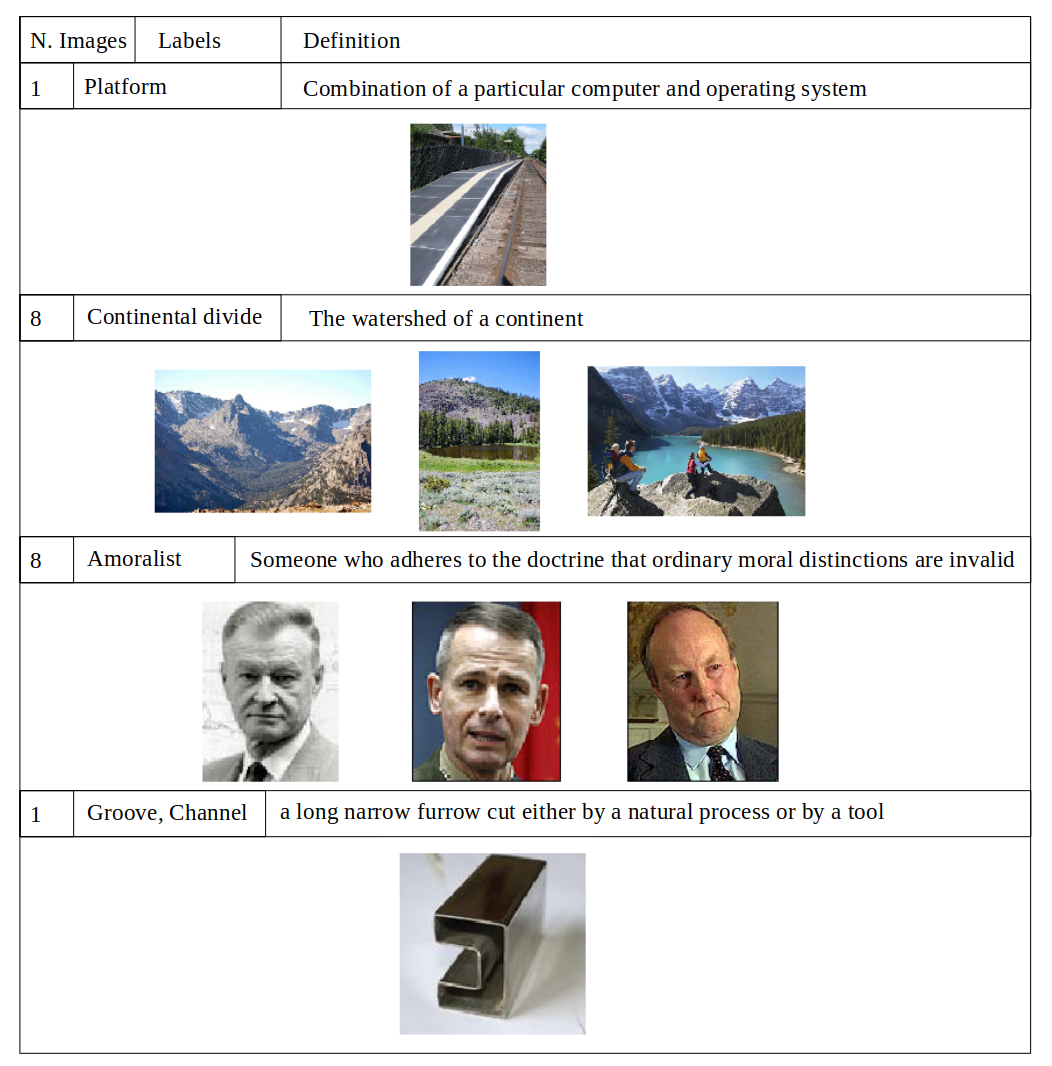}
\caption{
Examples of low sample population, visually ambiguous classes.
}
\end{figure}

\begin{figure}[h!]
\centering
\includegraphics[width=0.95\textwidth]{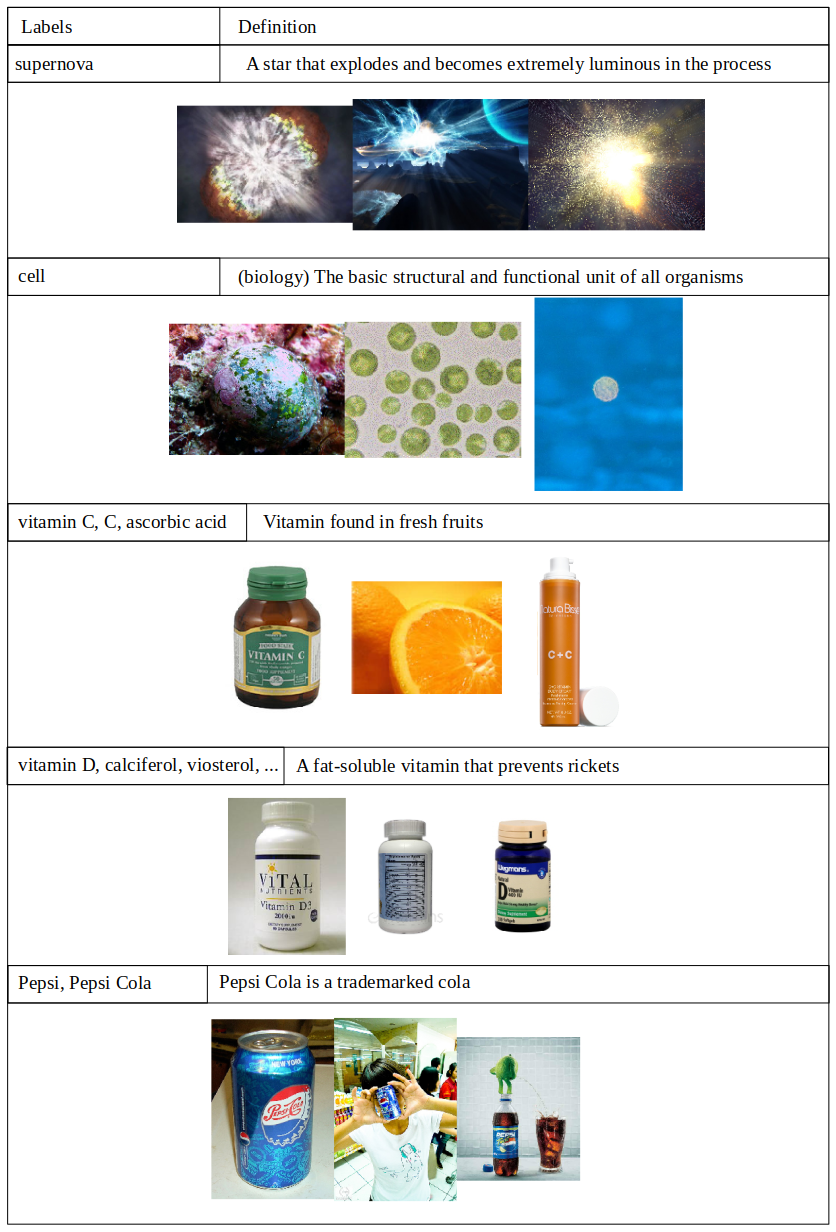}
\caption{
Examples of manually discarded classes. Cell and Supernova correspond to microscopic and astronomic scale images. 
Vitamin D, Vitamin C, and Pepsi were discarded as they require reading comprehension to identify.
}
\end{figure}

\end{document}